\documentclass[10pt,twocolumn,letterpaper]{article}
\usepackage[pagenumbers]{cvpr} 

\definecolor{cvprblue}{rgb}{0.21,0.49,0.74}
\usepackage[pagebackref,breaklinks,colorlinks,allcolors=cvprblue]{hyperref}
\usepackage{capt-of}
\usepackage{booktabs}
\usepackage{multirow}
\usepackage{graphicx}
\usepackage{xcolor}
\usepackage{colortbl}
\usepackage{algorithm}
\usepackage{algpseudocode}
\usepackage{tabularx}

\usepackage[most]{tcolorbox}

\newtcbox{\seqname}{
    on line,
    colback=black!8,
    colframe=black!8,
    boxrule=0pt,
    arc=2pt,
    boxsep=0pt,
    left=3pt,
    right=3pt,
    top=1pt,
    bottom=1pt
}

\definecolor{best}{HTML}{335F87}
\definecolor{second}{HTML}{805F72}
\definecolor{bestbg}{HTML}{EBF1F6}
\definecolor{secondbg}{HTML}{F1EAF0}
\definecolor{darkgreen}{RGB}{29,114,72}
\definecolor{darkblue}{RGB}{25,75,145}

\def\confName{CVPR}
\def\confYear{2027}

\title{Recall4D: Forgetting, Not Misalignment, \\
Limits Long-Horizon Feed-Forward 4D Reconstruction}

\author{Tianle Lyu\\
Carnegie Mellon University
}

\makeatletter
\g@addto@macro\@maketitle{\noindent\begin{minipage}{\textwidth}
    \captionsetup{hypcap=true,type=figure}
    \centering
    \resizebox{\linewidth}{!}{%
        \includegraphics[height=1in]{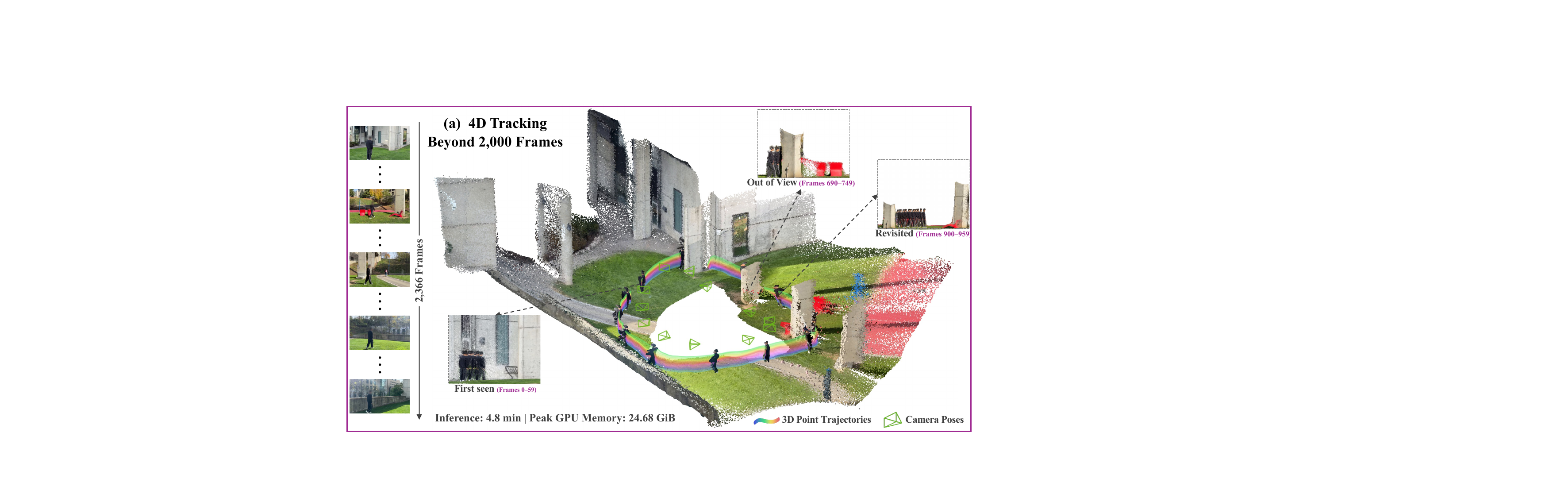}%
        \includegraphics[height=1in]{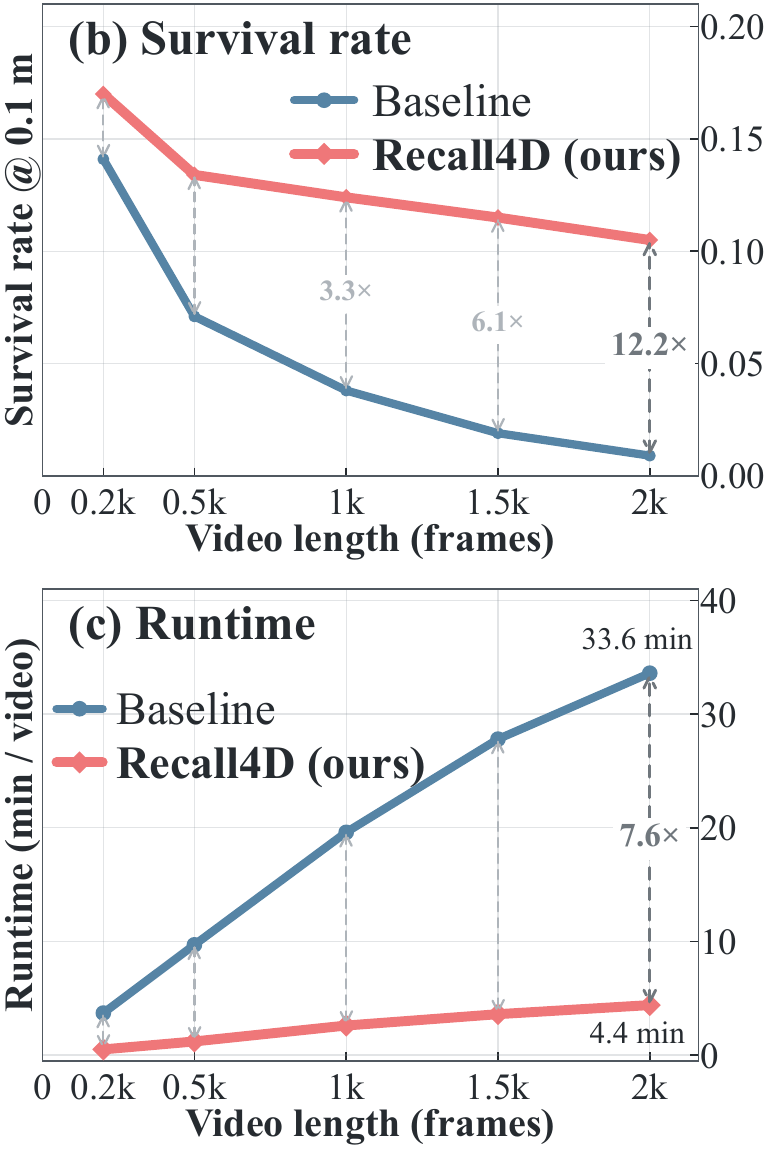}%
    }
    \vspace{-1.7em}
    \begingroup
    \caption{\textbf{(a)} To the best of our knowledge, Recall4D is the first method to perform 4D tracking over 2,000 frames in under 5 minutes on average, using less than 25 GiB of peak GPU memory while maintaining high accuracy. \textbf{(b)} Tracking survival rate at a 0.1 m threshold across increasing video lengths on PointOdyssey. \textbf{(c)} Recall4D tracks 2,000 frames in 4.4 minutes, $7.6\times$ faster than the chained baseline.}
    \label{fig1}
    \endgroup
    \par\vspace{12pt}
\end{minipage}\par}
\patchcmd{\maketitlesupplementary}{\vspace{1.0em}}{%
  \vspace{1.0em}\normalsize\noindent\begin{minipage}{\textwidth}
    \captionsetup{hypcap=true,type=figure}
    \centering
    \includegraphics[width=\linewidth]{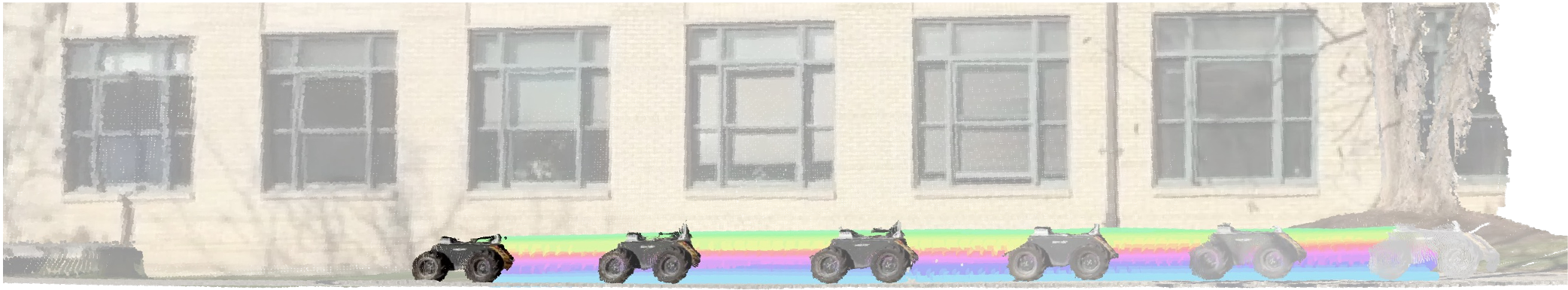}
    \vspace{-2em}
\caption{\textbf{Recall4D on a real-world video.} The static scene is reconstructed once and kept in the persistent world state, while the moving robot is tracked through sparse anchors. It is drawn at several time steps, with its trajectories in color. More results are in Sec.~\ref{secXE}.}
    \label{fig7}
    \par\vspace{12pt}
\end{minipage}\par}{}{%
  \PackageError{paper}{Could not add the supplementary teaser}{}%
}
\makeatother

\begin{document}
\maketitle
\begingroup
\setlength{\parskip}{0pt}
\setlength{\topskip}{10pt}
\patchcmd{\abstract}{\vspace*{12pt}}{\vspace*{12pt plus 12pt}}{}{%
  \PackageError{paper}{Could not adjust abstract heading spacing}{}%
}
\begin{abstract}
Feed-forward 4D trackers recover camera motion, depth, and point trajectories from short video clips, but chaining windows over long videos degrades their accuracy. We find that replacing inter-window alignment with ground-truth transforms leaves tracking error essentially unchanged. Static-point error instead grows with time since last observation: repeatedly predicting a static point from windows that no longer observe it corrupts previously recovered geometry. We call this failure forgetting. Merely retaining the last observed positions of unseen points reduces error by 26\% on PointOdyssey without modifying the tracker. We introduce Recall4D, a training-free framework that augments a frozen tracker with a persistent world state and sparse motion anchors. Static points retain their world-space positions outside the camera view and are refined by averaging observations decoded at sparse keyframes while in view. For dynamic objects, the tracker decodes sparse representative anchors, while geometric skinning propagates their displacements to dense queries. Across two long-horizon benchmarks, Recall4D improves the accuracy of its backbone while running 7.6--9.0$\times$ faster, and its accuracy gains carry over to five further backbones and three other long-sequence tasks, demonstrating its generality. Project page: \href{https://kincin.github.io/Recall4D/}{https://kincin.github.io/Recall4D/}.
\end{abstract}
\par
\endgroup

\section{Introduction}
\label{sec1}
Feed-forward models have changed how we recover 3D motion from video~\cite{wang2024dust3r,wang2025vggt,zhang2025monst3r}. Given a short clip, a single network pass now predicts camera poses, depth, and the 3D trajectories of query points, without per-video optimization~\cite{liu2026trace,xiao2025spatialtrackerv2,karhade2026any4d}. Real videos, however, are long. To process them, these models slide a window over the video and chain the windows: each window is aligned to the previous one, and every tracked point is handed over as a 3D position and a local descriptor~\cite{harley2022particle,karaev2024cotracker,jeon2026point4d}. This works for a few hundred frames. Then it breaks down: as Fig.~\ref{fig1}(b) shows, the survival rate of Point4D drops sharply as videos grow longer.

\begin{figure}[t]
    \centering

    \includegraphics[width=\columnwidth]{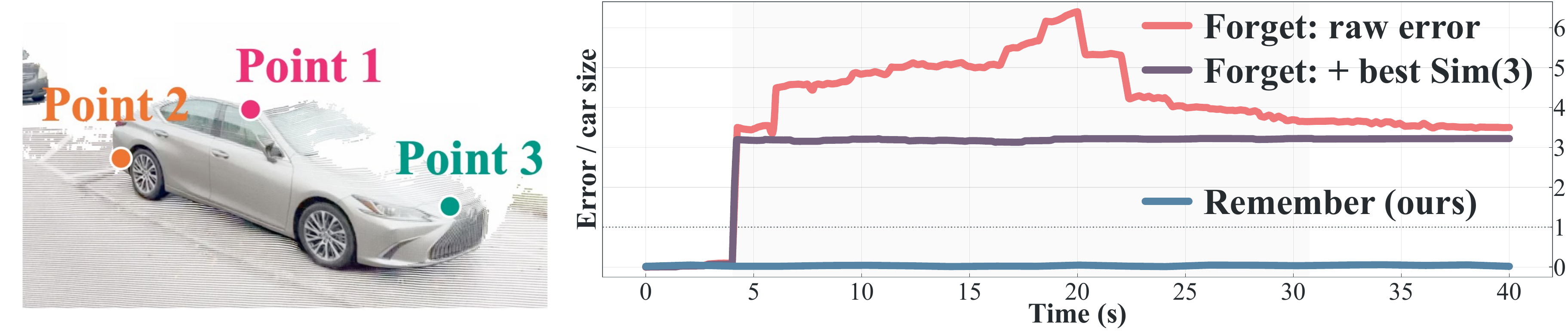}
    \par\vspace{-0em}
    \includegraphics[width=\columnwidth]{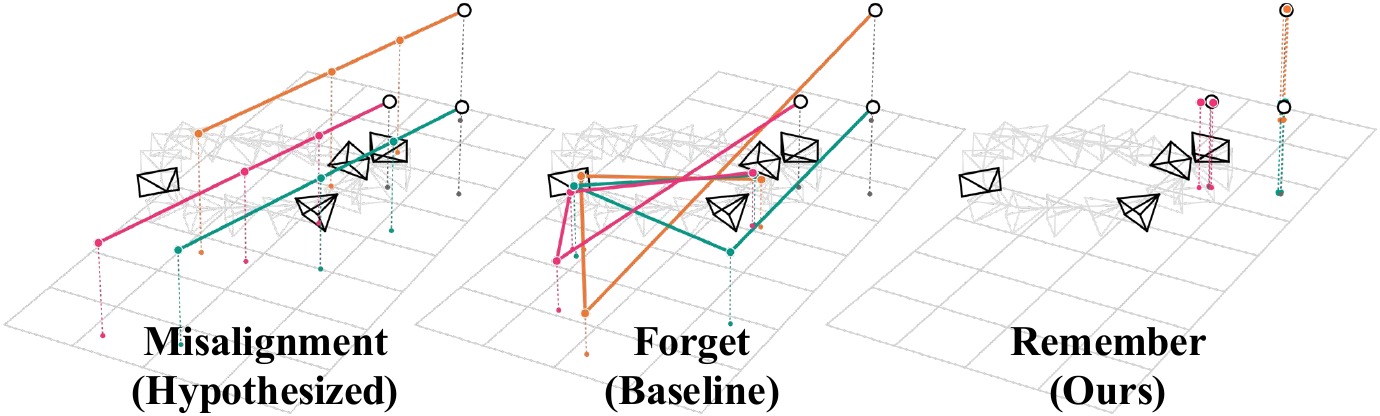}
    \vspace{-1.9em}
    \caption{\textbf{Forgetting, not misalignment.} Static points on a parked car: misalignment (left) would be undone by one Sim(3), but unseen points scatter independently (middle) and stay wrong after the best Sim(3) (top). Recall4D holds them in place (right).}
    \vspace{-1em}
    \label{fig2}
\end{figure}

The natural explanation is accumulated drift across window handovers, suggesting familiar remedies such as better alignment or loop closure~\cite{deng2026vggt,maggio2026vggt,sandstrom2025splat}. However, our analysis in Sec.~\ref{sec3} reveals a different dominant failure: \textit{forgetting rather than registration drift}. Replacing inter-window alignment with ground truth leaves the error unchanged. Instead, the error grows with how long a point has been out of view. Once a point leaves the view, the stateless tracker repeatedly re-predicts its position from windows that no longer observe it, causing even static points to wander over time, as illustrated in Fig.~\ref{fig2}. Simply retaining their last observed positions reduces this error by 26\% without modifying the model. These findings suggest that the key to long-range tracking is not better alignment, but a persistent memory of previously observed geometry~\cite{wang20253d,wang2025continuous,wu2026point3r}. This failure also exposes a deeper inefficiency: the tracker spends much of its computation predicting what it has already seen. Static points are decoded even after hundreds of frames out of view, although they never move.

For moving objects, the redundancy is equally fundamental: their points move, but not independently. Their trajectories are coupled by shared object motion~\cite{karaev2024cotracker,wang2025shape}, yet the tracker predicts each point as if its motion were independent. \textit{If points on the same object move together, why decode every one of them?} 
A few representative anchors can capture the shared motion and propagate it to the rest. The problem is therefore twofold: re-predicting what should be remembered, and decoding what can be inferred.

This diagnosis calls for a different fix: not a better alignment, but a memory; not more queries, but fewer. We introduce \textbf{Recall4D}, which equips a frozen feed-forward tracker with a persistent world state and sparse motion anchors. Static background points are written into the state in world coordinates. Whether a stored point lies within the camera view is decided by projecting the state, not the latest prediction, into the predicted camera. The state is refined from sparse keyframe observations and preserved when a point is out of view. At every frame, the state, rather than a fresh prediction, is the output. Moving objects are handled differently. A small set of anchors is tracked by the decoder, while the remaining points follow their motion through geometric skinning. The result is a stateful tracker that remembers static geometry and propagates dynamic motion without decoding every point at every frame.

Our main contributions are summarized as follows:
\begin{itemize}
  \item We diagnose forgetting of previously observed geometry, rather than inter-window registration drift, as a major cause of long-horizon degradation in feed-forward 4D reconstruction, supported by a fine-grained error analysis.
  \item We propose Recall4D, a training-free framework that equips frozen feed-forward 4D trackers with a persistent world state for static geometry and sparse motion anchors for dynamic objects, avoiding redundant predictions through geometric skinning.
  \item We demonstrate that Recall4D consistently improves long-horizon tracking accuracy and computational efficiency across multiple datasets, backbones, and tasks.
\end{itemize}
 
\section{Related Work}
\label{sec2}
\noindent\textbf{Long-range 2D and 3D point tracking.}
TAP-Vid~\cite{doersch2022tap} established point tracking as a benchmark task, followed by window-based 2D trackers~\cite{harley2022particle,doersch2023tapir,doersch2024bootstap,karaev2024cotracker,karaev2025cotracker3}. SpatialTracker~\cite{xiao2024spatialtracker}, SpatialTrackerV2~\cite{xiao2025spatialtrackerv2}, TAPIP3D~\cite{zhang2026tapip3d}, and DELTA~\cite{ngo2025delta} extend tracking to 3D with depth, camera motion, or persistent geometry, and long-range benchmarks~\cite{zheng2023,karaev2023dynamicstereo,koppula2024tapvid} enable systematic evaluation. Concurrently, TrackEverything~\cite{jain2026trackeverything} achieves long-horizon tracking through voxel-based deduplication and classification of points as static or dynamic. We share this split but, instead of building a new tracker, diagnose why existing trackers fail over long videos and repair them without training.

\noindent\textbf{Feed-forward 3D and 4D reconstruction.}
Feed-forward reconstruction has advanced from pairwise pointmap prediction~\cite{wang2024dust3r,leroy2024grounding} to multi-view geometry estimation~\cite{wang2025vggt,wang2026pi} and dynamic reconstruction~\cite{zhang2025monst3r}. Recent methods directly predict 4D geometry and point trajectories~\cite{feng2025st4rtrack,karhade2026any4d,zhang2026efficiently,sucar2026v,liu2026trace,luo20264rc,lu2026track4world}, including our backbone Point4D~\cite{jeon2026point4d}. Extending such models to long videos requires chaining overlapping windows through alignment, sometimes with loop closure~\cite{deng2026vggt}. These pipelines treat alignment as the bottleneck; we show that, for point tracking, re-predicting previously observed geometry matters more.
\begin{table*}[t]
\centering
\setlength{\tabcolsep}{4pt}
\caption{\textbf{Where the long-video error comes from.} EPE by point class on PointOdyssey~\cite{zheng2023} val (15 sequences, 1,000 frames); APD and SR are reported over all retained points. GT-pose alignment uses oracle inter-window transforms; hold unseen retains the last observed position after 40 unseen frames. Scene-exiting GT tracks are excluded for all methods.}
\vspace{-0.8em}
\label{tab1}
\resizebox{\linewidth}{!}{\begin{tabular}{c l ccc ccc ccc}
\toprule
 &  & \multicolumn{3}{c}{All points} & \multicolumn{3}{c}{Static points} & \multicolumn{3}{c}{Dynamic points} \\
\cmidrule(lr){3-5}\cmidrule(lr){6-8}\cmidrule(lr){9-11}
 &  & EPE$\downarrow$ & APD$\uparrow$ & SR$\uparrow$ & visible & occluded & out of frame & visible & occluded & out of frame \\
\midrule
 & Share of (point, frame) pairs & 100\% & -- & -- & 8.4\% & 2.4\% & 17.0\% & 40.2\% & 17.9\% & 14.2\% \\
\midrule
 & Chained baseline & 1.569 & 0.281 & 0.193 & 1.246 & 1.163 & 1.799 & 1.193 & 1.280 & 1.856 \\

(a) & \quad + GT-pose alignment (no registration drift) & 1.568 {\scriptsize(0\%)} & 0.288 {\scriptsize(+2\%)} & 0.199 {\scriptsize(+3\%)} & 1.252 {\scriptsize(+0\%)} & 1.190 {\scriptsize(+2\%)} & 1.878 {\scriptsize(+4\%)} & 1.163 {\scriptsize($-$3\%)} & 1.243 {\scriptsize($-$3\%)} & 1.876 {\scriptsize(+1\%)} \\

(b) & \quad + hold unseen points (predicted view) & 1.168 {\scriptsize($-$26\%)} & 0.384 {\scriptsize(+37\%)} & 0.260 {\scriptsize(+35\%)} & 1.029 {\scriptsize($-$17\%)} & 1.024 {\scriptsize($-$12\%)} & 1.115 {\scriptsize($-$38\%)} & 1.112 {\scriptsize($-$7\%)} & 1.196 {\scriptsize($-$7\%)} & 1.629 {\scriptsize($-$12\%)} \\

(c) & \quad + GT-pose alignment + hold unseen points & 1.173 {\scriptsize($-$25\%)} & 0.391 {\scriptsize(+39\%)} & 0.260 {\scriptsize(+35\%)} & 1.032 {\scriptsize($-$17\%)} & 1.030 {\scriptsize($-$11\%)} & 1.133 {\scriptsize($-$37\%)} & 1.094 {\scriptsize($-$8\%)} & 1.172 {\scriptsize($-$8\%)} & 1.658 {\scriptsize($-$11\%)} \\

(d) & \quad + hold unseen points (GT visibility) & 1.043 {\scriptsize($-$33\%)} & 0.436 {\scriptsize(+55\%)} & 0.282 {\scriptsize(+46\%)} & 0.971 {\scriptsize($-$22\%)} & 0.897 {\scriptsize($-$23\%)} & 0.768 {\scriptsize($-$57\%)} & 1.060 {\scriptsize($-$11\%)} & 1.165 {\scriptsize($-$9\%)} & 1.576 {\scriptsize($-$15\%)} \\
\midrule
\rowcolor{gray!15}
 & \textbf{Recall4D (Ours)} & \textbf{0.889} {\scriptsize($-$43\%)} & \textbf{0.471} {\scriptsize(+68\%)} & \textbf{0.359} {\scriptsize(+86\%)} & \textbf{0.434} {\scriptsize($-$65\%)} & \textbf{0.497} {\scriptsize($-$57\%)} & \textbf{0.433} {\scriptsize($-$76\%)} & \textbf{0.948} {\scriptsize($-$21\%)} & \textbf{1.064} {\scriptsize($-$17\%)} & \textbf{1.534} {\scriptsize($-$17\%)} \\
\bottomrule
\end{tabular}}
\vspace{-1em}
\end{table*}

\noindent\mbox{\textbf{State and sparse control for long videos.}}
Streaming reconstruction models retain temporal information through spatial memory~\cite{wang20253d,wu2026point3r}, recurrent states~\cite{wang2025continuous}, or cached tokens~\cite{zhuo2026streaming}, with confidence-guided~\cite{chen2026ttt3r}, adaptive~\cite{zheng2026ttsa3r}, and ray-aware~\cite{li2026ray} state updates, while keyframe-based SLAM~\cite{sandstrom2025splat,maggio2026vggt} maintains persistent maps. Window-chained trackers, in contrast, keep no state at all; we attach an explicit, parameter-free world state to a frozen one, where simple averaging proves sufficient. For dynamic objects, sparse control points have long supported non-rigid deformation~\cite{sumner2007embedded,newcombe2015dynamicfusion} and dynamic Gaussian representations~\cite{huang2024sc,wang2025shape} through per-scene optimization. Our anchors are instead decoded directly by the frozen tracker, with geometric skinning propagating their motion to dense points.
\section{Diagnosing Long-Horizon Degradation}
\label{sec3}

\begin{figure}[t]
    \centering
    \includegraphics[width=\columnwidth]{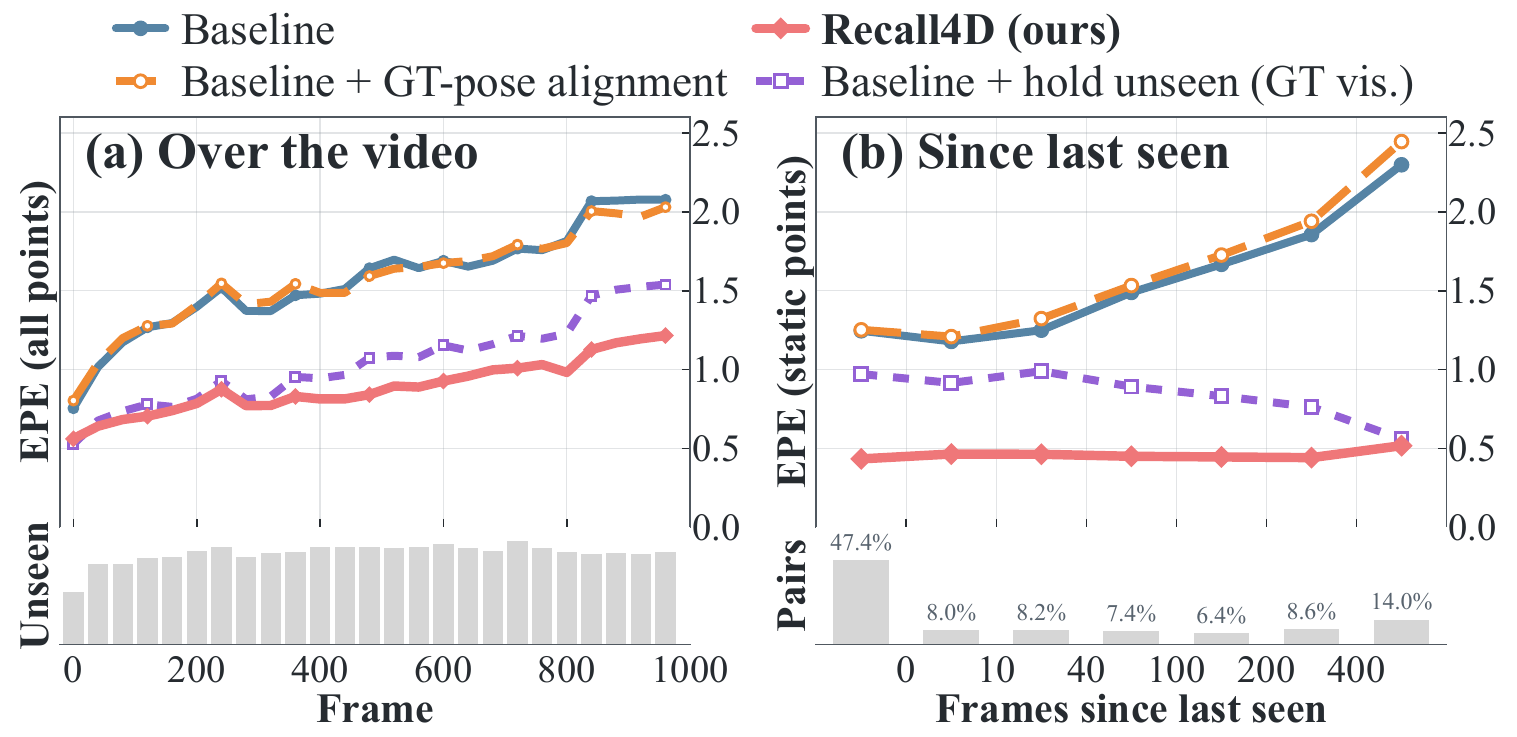}
    \vspace{-1.8em}
    \caption{\textbf{Error grows with time since last seen.} (a) EPE over time. (b) Static-point EPE versus frames since last seen. GT-pose alignment matches the baseline; holding unseen points prevents the rise. Bars show unseen fractions (a) and pair fractions (b).}
    \vspace{-1.5em}
    \label{fig3}
\end{figure}
\begin{figure*}[!t]
    \centering
    \resizebox{\textwidth}{!}{\includegraphics{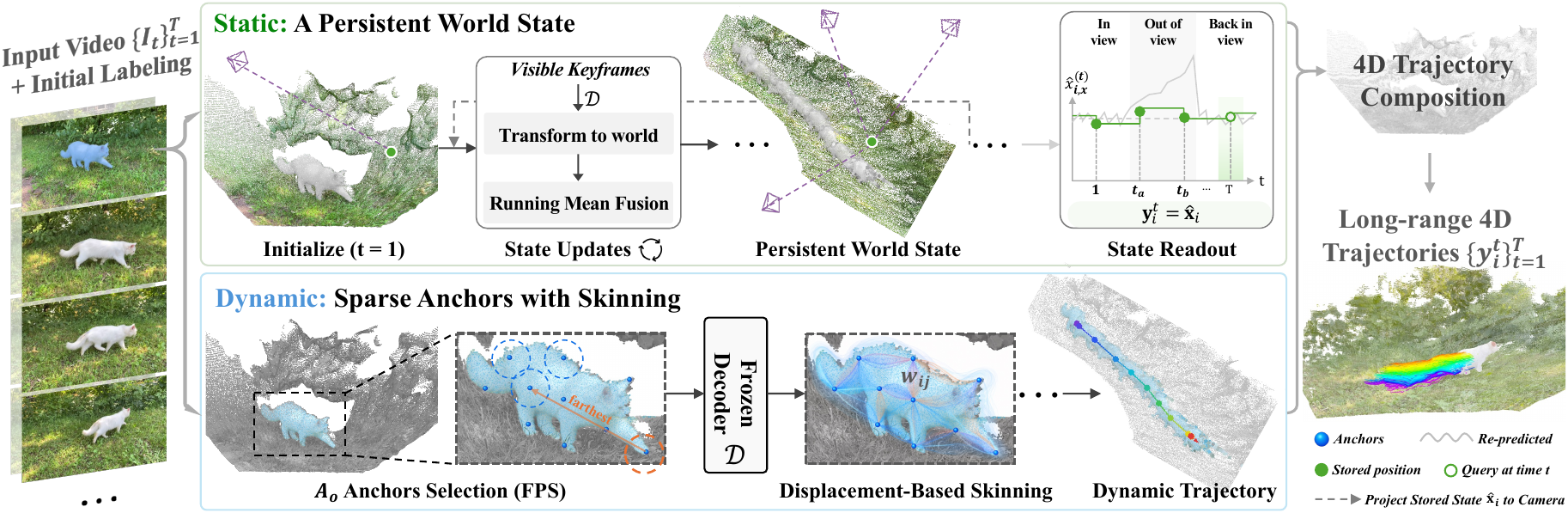}}
    \vspace{-2em}
    \caption{\textbf{Overview of Recall4D.} Recall4D maintains a persistent world state to preserve background geometry, refining stored positions through running-mean fusion at in-view keyframes. For moving objects, a frozen decoder $\mathcal{D}$ tracks only sparse anchors, whose displacements propagate to dense queries through geometric skinning. Both are composed into long-range 4D trajectories $\{\mathbf{y}_i^t\}_{t=1}^{T}$.}
    \vspace{-1em}
    \label{fig4}
\end{figure*}
\noindent Why do feed-forward 4D trackers fail as videos grow longer? The natural explanation is accumulated drift: small misalignments at each window handover compound and  corrupt the reconstruction. We first pursued this hypothesis by improving cross-window alignment and smoothing predictions over the overlapping frames, but neither helped, as detailed in Supp.~\ref{secXA}. The problem therefore seems to lie not in how windows are connected, but in how point estimates are carried across them. To find out, we evaluate the chained baseline on the 1,000-frame PointOdyssey~\cite{zheng2023} validation set. Tab.~\ref{tab1} breaks the error down by point class, and Fig.~\ref{fig3} shows how it evolves over time.

\noindent\textbf{Is alignment drift really the culprit?}
At each window handover, the baseline estimates a Sim(3) from the predicted depth of the overlapping frames. Relative to the ground-truth camera poses, these estimates are off by $1.2^\circ$ in rotation and $3.4\%$ in scale on average. In Tab.~\ref{tab1}(a), we replace every one of them with the transform implied by the ground-truth poses, so that the chain carries no registration error. If accumulated misalignment drove the degradation, this oracle would recover much of the lost accuracy. \textbf{\textit{It does not.}} EPE stays at 1.568 against 1.569 for the baseline, and the error curve in Fig.~\ref{fig3}(a) is indistinguishable from the original over the whole video. Whatever degrades Point4D over time, it is not how its windows are registered.

\noindent\textbf{A static point does not move. Its prediction does.}
In Fig.~\ref{fig3}(b), we therefore plot the error against the time since a point was last observed. For static points, EPE rises from 1.25 while visible to 2.30 after more than 400 unseen frames, and with perfect registration it still reaches 2.45. The ground-truth position of a static point never changes; what changes is the tracker's estimate. Once the point leaves the view, Point4D keeps decoding its position from windows that contain no observation of it. Each new estimate replaces the previous one and is handed to the next window, progressively corrupting geometry that was already recovered. The same holds, to a lesser degree, for points in view: every window re-derives a point from the single 3D position handed over to it, and nothing ties the new estimate to the earlier ones. The EPE of visible static points thus doubles from 0.83 in the first 200 frames to 1.71 in the last 200, although the share of unseen points stays near 50\% after the first window. The tracker is not losing the scene because the scene moves or because its windows drift apart, but because it keeps re-predicting what it should have remembered. We call this failure
\textit{forgetting}.

\noindent\textbf{What if we simply stop re-predicting?}
The diagnosis suggests a direct test. Without modifying the network, improving alignment, or adding a learned memory, we change one rule: once a point has been unseen for a whole window, it keeps its last observed position until it is seen again. In Tab.~\ref{tab1}(b), visibility is decided by Point4D's own cameras, which detect points leaving the frame but not occlusions. This rule alone reduces EPE by 26\% and raises SR from 0.193 to 0.260. With ground-truth visibility in Tab.~\ref{tab1}(d), the reduction reaches 33\%. Adding GT-pose alignment on top of the hold in Tab.~\ref{tab1}(c) changes nothing. Remembering observed positions thus removes a quarter to a third of the error, whereas perfect registration removes none. Forgetting, not registration drift, drives long-horizon degradation.

\noindent\textbf{Not all points need to be treated alike.}
Tab.~\ref{tab1} reveals a second insight. With ground-truth visibility, holding cuts the out-of-frame EPE of static points by 57\%, but the EPE of dynamic points by only 9\% to 15\%: their positions must keep evolving, and a frozen position is only marginally better than a forgotten one. Static points are also as inaccurate as dynamic ones even while visible, with an EPE of 1.25 against 1.19, despite having no true motion: decoding the same static point anew in every window injects fresh prediction noise. Static points should therefore be stored and refined over repeated observations rather than re-predicted. Dynamic points pose a different problem: their error depends on backbone motion prediction, which holding cannot improve, and decoding them independently is wasteful.

% These findings point to a different strategy for long-range tracking: \textbf{\textit{1)}} static geometry should persist across windows and be refined when re-observed, and \textbf{\textit{2)}} dynamic motion should be decoded sparsely and propagated to the remaining points.
\section{Method}
\label{sec4}

Motivated by the findings in Sec.~\ref{sec3}, we propose Recall4D, as illustrated in Fig.~\ref{fig4}. Given a monocular video $\{I_t\}_{t=1}^{T}$ and $N$ query points with initial pixel coordinates $\{\mathbf{u}_i\}_{i=1}^{N}$ in the first frame, our goal is to estimate their long-horizon 3D trajectories $\{\mathbf{y}_i^t\}_{t=1}^{T}$ in a common world coordinate system.
\subsection{Preliminaries}
\label{sec41}
Our framework targets window-chained feed-forward 4D trackers, which process long videos through overlapping temporal windows $C_k$, indexed by $k$, using an encoder $\mathcal{E}$ and a query decoder $\mathcal{D}$. Each query $i$ consists of a 3D position $\mathbf{q}_i^{(k)}$ in the window's reference camera coordinates and an appearance descriptor $\phi_i$ extracted from the query frame. The decoder predicts its 3D position at each frame $t \in C_k$:
\begin{equation}
    \mathbf{x}_{i,k}^{t}
    = \mathcal{D}\left(\mathcal{E}(C_k);\, \mathbf{q}_i^{(k)}, \phi_i\right)_t .
    \label{eq1}
\end{equation}
Here, $\mathbf{x}_{i,k}^{t}$ denotes the predicted 3D position of query $i$ at frame $t$ from window $k$, expressed in its reference coordinates, and $\mathcal{D}(\cdot)_t$ denotes the positional output at frame $t$.

The backbone also predicts per-frame depth maps and camera parameters $(R_t, \mathbf{c}_t, K_t)$, denoting camera-to-world rotation, camera center, and intrinsics, respectively. Overlapping windows are aligned through similarity transformations, where $G_k \in \mathrm{Sim}(3)$ maps world coordinates to the reference coordinates of window $k$. Accordingly, $G_k^{-1}$ transforms decoded positions back to the world frame. Camera parameters are expressed in the common world frame after alignment.

Across windows, queries are initialized from previous predictions. Let $t_k$ denote the first frame of window $C_k$, which overlaps with the preceding window. The query is transferred into the new reference frame as
\begin{equation}
    \mathbf{q}_i^{(k)}
    = \left(G_k \circ G_{k-1}^{-1}\right)\left(\mathbf{x}_{i,k-1}^{t_k}\right),
    \qquad
    \phi_i^{(k)} = \phi_i .
    \label{eq2}
\end{equation}
Here, $\mathbf{x}_{i,k-1}^{t_k}$ is the previous window's prediction at frame $t_k$, and $\circ$ denotes transformation composition. Predictions are blended over overlapping frames, but no persistent history of past 3D observations is maintained.

As shown in Sec.~\ref{sec3}, repeatedly predicting unseen static points corrupts recovered geometry, while improved inter-window alignment offers little benefit. This motivates preserving static geometry and refining it upon re-observation. Simultaneously, correlated motion among dynamic points makes independent decoding redundant. Accordingly, Recall4D maintains a persistent world state for static geometry in Sec.~\ref{sec42} and propagates dynamic motion through sparse anchors with geometric skinning in Sec.~\ref{sec43}.
\subsection{A Persistent World State}
\label{sec42}
We introduce a training-free persistent world state to maintain temporally consistent 3D positions for background points. We obtain per-frame foreground masks $M_t$ using an off-the-shelf segmentation model, where $M_t=1$ indicates foreground. Queries outside the foreground mask in the first frame are treated as static:
\begin{equation}
    \mathcal{B} = \left\{ i : M_1(\mathbf{u}_i) = 0 \right\}.
    \label{eq3}
\end{equation}
Each $i \in \mathcal{B}$ maintains a persistent state $s_i=(\hat{\mathbf{x}}_i, n_i)$, consisting of its world-space position and the number of fused observations. The state is initialized by lifting the query to 3D using predicted depth, with $n_i=0$, so the first observation replaces this estimate.

We determine whether a stored point lies within the camera view by projecting its world-space position into the predicted camera at frame $t$:
\begin{equation}
\begin{aligned}
    \mathbf{p}_i^t &= R_t^\top\left(\hat{\mathbf{x}}_i - \mathbf{c}_t\right),\\
    v_i^t &= \mathbb{I}\left[(\mathbf{p}_i^t)_z > 0 \,\land\, \pi(K_t\mathbf{p}_i^t) \in \Omega\right].
\end{aligned}
\label{eq4}
\end{equation}
Here, $\pi$ denotes perspective division, $\Omega$ is the image domain, and $\mathbb{I}$ is the indicator function. The indicator $v_i^t$ tests whether a point lies within the camera frustum, without resolving occlusion. Crucially, this decision uses the persistent world state rather than the decoder's latest prediction.

In each window $k$, we update the persistent state at $m$ uniformly sampled keyframes $\mathcal{K}_k$ from the newly introduced frames, excluding the query frame. At each keyframe $\tau \in \mathcal{K}_k$, background points satisfying $v_i^\tau=1$ are decoded using their stored positions as 3D queries, yielding new observations in world coordinates:
\begin{equation}
    \mathbf{x}_i^\tau
    = G_k^{-1}\left(\mathcal{D}\left(\mathcal{E}(C_k);\, G_k(\hat{\mathbf{x}}_i), \phi_i\right)_\tau\right).
    \label{eq5}
\end{equation}
Here, $\mathbf{x}_i^\tau$ is the decoded observation expressed in the common world coordinate system. These observations are fused into the persistent state via a running mean:
\begin{equation}
    \hat{\mathbf{x}}_i \leftarrow \frac{n_i\hat{\mathbf{x}}_i + \mathbf{x}_i^\tau}{n_i+1},
    \qquad
    n_i \leftarrow n_i + 1 .
    \label{eq6}
\end{equation}
Points outside the camera view retain their stored positions without further updates.

At each frame, the output for each background query is directly retrieved from its current persistent state as $\mathbf{y}_i^t=\hat{\mathbf{x}}_i(t)$, for all $i\in\mathcal{B}$ and $t=1,\ldots,T$. Here, $\hat{\mathbf{x}}_i(t)$ denotes the stored position used to produce the output at frame $t$. Rather than repeatedly predicting background geometry at every frame, our approach preserves out-of-view geometry and refines in-view estimates through sparse keyframe observations. This avoids redundant background decoding while maintaining temporally consistent positions.

We adopt uniform averaging, as more elaborate fusion strategies provide no consistent improvement in our experiments, with residual errors primarily arising from systematic bias along the viewing ray.
\subsection{Sparse Anchors with Skinning}
\label{sec43}
A key observation is that points belonging to the same moving object often exhibit correlated motion, rather than moving independently. This shared motion structure makes dense, independent trajectory decoding computationally redundant. We exploit this redundancy by tracking sparse representative anchors and propagating their motion to dense queries through geometric skinning.

For each foreground instance $o$ identified in the first-frame mask, we lift its mask pixels to 3D using predicted depth and transform them into world coordinates, forming a point set $X_o$. We select $A_o$ representative anchors through farthest point sampling:
\begin{equation}
    \left\{\bar{\mathbf{a}}_j^{(0)}\right\}_{j=1}^{A_o} = \mathrm{FPS}\left(X_o, A_o\right).
    \label{eq7}
\end{equation}
Here, $o$ indexes foreground instances, $j$ indexes anchors, and $A_o$ denotes the number of anchors per instance. Each anchor is tracked by the frozen decoder $\mathcal{D}$, producing its world-space position $\mathbf{a}_j^t$ at every frame $t \in C_k$.

Rather than decoding every foreground query independently, we approximate its motion from nearby anchors. Each query $i$, with initial world-space position $\bar{\mathbf{x}}_i^{(0)}$, is associated with its $H$ nearest anchors $\mathcal{N}(i)$ from the same instance. We compute distance-based skinning weights:
\begin{equation}
\begin{aligned}
    w_{ij} &= \frac{\exp\left[-(d_{ij}/\lambda_i)^2\right]}
                  {\sum_{\ell\in\mathcal{N}(i)} \exp\left[-(d_{i\ell}/\lambda_i)^2\right]},\\
    d_{ij} &= \left\|\bar{\mathbf{x}}_i^{(0)} - \bar{\mathbf{a}}_j^{(0)}\right\|_2,
    \qquad
    \lambda_i = 2\min_{\ell\in\mathcal{N}(i)} d_{i\ell}.
\end{aligned}
\label{eq8}
\end{equation}
Here, $H$ is the number of nearest anchors, $d_{ij}$ is the initial distance between query $i$ and anchor $j$, and $\lambda_i$ controls the spatial decay of the weights. Closer anchors receive larger weights. At each frame $t\in C_k$, the output of each foreground query $i\notin\mathcal{B}$ blends the displacements of its neighboring anchors:
\begin{equation}
    \mathbf{y}_i^t
    = \bar{\mathbf{x}}_i^{(k)}
    + \sum_{j\in\mathcal{N}(i)} w_{ij}\left(\mathbf{a}_j^t - \bar{\mathbf{a}}_j^{(k)}\right).
    \label{eq9}
\end{equation}
Here, $\bar{\mathbf{x}}_i^{(k)}$ and $\bar{\mathbf{a}}_j^{(k)}$ denote the query and anchor rest positions at the beginning of window $k$, respectively. At each window transition, these rest positions are updated from the previous predictions, while the anchor associations and skinning weights remain unchanged. Consequently, only $A_o$ anchors per instance require trajectory decoding, while the remaining foreground queries are reconstructed through lightweight geometric skinning.
\section{Experimental Setup}
\label{sec5}
\begin{table*}[t]
\centering
\caption{Cross-dataset comparison on PointOdyssey test~\cite{zheng2023} and Dynamic Replica test~\cite{karaev2023dynamicstereo}. Results are reported for 200- and 500-frame sequences. \colorbox{bestbg}{\textcolor{best}{\textbf{Best}}} and \colorbox{secondbg}{\textcolor{second}{\textbf{second-best}}} results are highlighted accordingly.}
\vspace{-0.85em}
\label{tab2}
\resizebox{\textwidth}{!}{
\begin{tabular}{lcccccccc|cccccccc}
\toprule
\multirow{2}{*}{Method}
& \multicolumn{4}{c}{PointOdyssey (200 frames)}
& \multicolumn{4}{c}{PointOdyssey (500 frames)}
& \multicolumn{4}{c}{Dynamic Replica (200 frames)}
& \multicolumn{4}{c}{Dynamic Replica (500 frames)} \\
\cmidrule(lr){2-5}
\cmidrule(lr){6-9}
\cmidrule(lr){10-13}
\cmidrule(lr){14-17}
& {\small EPE$\downarrow$} & {\small APD$\uparrow$} & {\small SR$\uparrow$} & {\small Time$\downarrow$}
& {\small EPE$\downarrow$} & {\small APD$\uparrow$} & {\small SR$\uparrow$} & {\small Time$\downarrow$}
& {\small EPE$\downarrow$} & {\small APD$\uparrow$} & {\small SR$\uparrow$} & {\small Time$\downarrow$}
& {\small EPE$\downarrow$} & {\small APD$\uparrow$} & {\small SR$\uparrow$} & {\small Time$\downarrow$} \\
\midrule
SpatialTrackV2~\cite{xiao2025spatialtrackerv2}
& \cellcolor{bestbg}\textcolor{best}{\textbf{0.492}} & 0.617 & 0.504 & 179.7
& 1.037 & \cellcolor{secondbg}\textcolor{second}{\textbf{0.504}} & 0.347 & 481.5
& \cellcolor{bestbg}\textcolor{best}{\textbf{0.236}} & \cellcolor{secondbg}\textcolor{second}{\textbf{0.754}} & 0.686 & 225.6
& \cellcolor{secondbg}\textcolor{second}{\textbf{0.263}} & 0.734 & 0.613 & 622.4 \\

TraceAnything~\cite{liu2026trace}
& 2.254 & 0.220 & 0.166 & \cellcolor{secondbg}\textcolor{second}{\textbf{31.4}}
& 2.572 & 0.137 & 0.072 & \cellcolor{secondbg}\textcolor{second}{\textbf{84.3}}
& 0.577 & 0.571 & 0.499 & \cellcolor{secondbg}\textcolor{second}{\textbf{36.8}}
& 0.685 & 0.508 & 0.400 & \cellcolor{secondbg}\textcolor{second}{\textbf{92.5}} \\

Any4D~\cite{karhade2026any4d}
& 0.857 & 0.473 & 0.391 & 69.2
& 1.473 & 0.296 & 0.209 & 198.3
& 0.329 & 0.712 & 0.657 & 69.5
& 0.433 & 0.667 & 0.578 & 207.4 \\

V-DPM~\cite{sucar2026v}
& 0.627 & 0.617 & 0.512 & 525.3
& 1.281 & 0.391 & 0.320 & 1377
& 0.368 & 0.648 & 0.595 & 524.2
& 0.501 & 0.576 & 0.491 & 1376 \\

4RC~\cite{luo20264rc}
& 0.780 & 0.570 & 0.508 & 58.7
& 1.626 & 0.347 & 0.283 & 179.8
& 0.295 & 0.759 & \cellcolor{secondbg}\textcolor{second}{\textbf{0.712}} & 59.1
& 0.444 & 0.657 & 0.580 & 169.4 \\

Track4World~\cite{lu2026track4world}
& 1.004 & 0.436 & 0.317 & 322.7
& 1.689 & 0.303 & 0.183 & 912.5
& 0.342 & 0.666 & 0.572 & 319.8
& 0.508 & 0.554 & 0.418 & 919.2 \\

TrackEverything$^\mathsf{d}$~\cite{jain2026trackeverything}
& 0.862 & \cellcolor{bestbg}\textcolor{best}{\textbf{0.656}} & \cellcolor{bestbg}\textcolor{best}{\textbf{0.582}} & 580.8
& OOM & OOM & OOM & OOM
& 0.277 & 0.727 & 0.695 & 142.3
& 0.287 & \cellcolor{secondbg}\textcolor{second}{\textbf{0.717}} & \cellcolor{secondbg}\textcolor{second}{\textbf{0.665}} & 391.5 \\

Point4D~\cite{jeon2026point4d}
& 0.661 & 0.564 & 0.489 & 226.4
& \cellcolor{secondbg}\textcolor{second}{\textbf{0.979}} & 0.466 & \cellcolor{secondbg}\textcolor{second}{\textbf{0.370}} & 582.1
& 0.255 & 0.751 & 0.710 & 291.2
& 0.307 & 0.700 & 0.646 & 753.2 \\
\midrule

\textbf{Recall4D (Ours)}
& \cellcolor{secondbg}\textcolor{second}{\textbf{0.609}} & \cellcolor{secondbg}\textcolor{second}{\textbf{0.620}} & \cellcolor{secondbg}\textcolor{second}{\textbf{0.528}} & \cellcolor{bestbg}\textcolor{best}{\textbf{28.8}}
& \cellcolor{bestbg}\textcolor{best}{\textbf{0.836}} & \cellcolor{bestbg}\textcolor{best}{\textbf{0.521}} & \cellcolor{bestbg}\textcolor{best}{\textbf{0.440}} & \cellcolor{bestbg}\textcolor{best}{\textbf{74.9}}
& \cellcolor{secondbg}\textcolor{second}{\textbf{0.244}} & \cellcolor{bestbg}\textcolor{best}{\textbf{0.757}} & \cellcolor{bestbg}\textcolor{best}{\textbf{0.725}} & \cellcolor{bestbg}\textcolor{best}{\textbf{32.8}}
& \cellcolor{bestbg}\textcolor{best}{\textbf{0.262}} & \cellcolor{bestbg}\textcolor{best}{\textbf{0.736}} & \cellcolor{bestbg}\textcolor{best}{\textbf{0.690}} & \cellcolor{bestbg}\textcolor{best}{\textbf{84.1}} \\

\bottomrule
\end{tabular}
}
\vspace{-0.5em}
\end{table*}
\begin{table*}[t]
\captionsetup{justification=raggedright,singlelinecheck=false}
\caption{Long-horizon comparison on PointOdyssey test~\cite{zheng2023}, with Recall4D applied to TrackEverything$^\mathsf{s}$~\cite{jain2026trackeverything} and Point4D~\cite{jeon2026point4d} at 1,000, 1,500, and 2,000 frames. \colorbox{bestbg}{\textcolor{best}{\textbf{Best}}} and \colorbox{secondbg}{\textcolor{second}{\textbf{second-best}}} results are highlighted accordingly.}
\centering
\vspace{-0.85em}
\label{tab3}
\resizebox{\textwidth}{!}{
\begin{tabular}{lccccccccccccccc}
\toprule
\multirow{2}{*}{Method}
& \multicolumn{5}{c}{1,000 frames}
& \multicolumn{5}{c}{1,500 frames}
& \multicolumn{5}{c}{2,000 frames} \\
\cmidrule(lr){2-6}
\cmidrule(lr){7-11}
\cmidrule(lr){12-16}
& {\small EPE$\downarrow$} & {\small EPE$_\text{dyn}\downarrow$} & {\small APD$\uparrow$} & {\small SR$\uparrow$} & {\small Time$\downarrow$}
& {\small EPE$\downarrow$} & {\small EPE$_\text{dyn}\downarrow$} & {\small APD$\uparrow$} & {\small SR$\uparrow$} & {\small Time$\downarrow$}
& {\small EPE$\downarrow$} & {\small EPE$_\text{dyn}\downarrow$} & {\small APD$\uparrow$} & {\small SR$\uparrow$} & {\small Time$\downarrow$} \\
\midrule

TrackEverything$^\mathsf{s}$~\cite{jain2026trackeverything}
& 2.577 & 2.812 & 
\cellcolor{secondbg}\textcolor{second}{\textbf{0.586}} & 
\cellcolor{secondbg}\textcolor{second}{\textbf{0.475}} & 534
& 4.924 & 5.379 & 
\cellcolor{secondbg}\textcolor{second}{\textbf{0.534}} & 
\cellcolor{secondbg}\textcolor{second}{\textbf{0.434}} & 748
& 4.984 & 5.352 & 
\cellcolor{secondbg}\textcolor{second}{\textbf{0.497}} & 
\cellcolor{secondbg}\textcolor{second}{\textbf{0.395}} & 852 \\

\quad \textbf{+ Ours}
& \cellcolor{secondbg}\textcolor{second}{\textbf{2.569}} & 2.799 
& \cellcolor{bestbg}\textcolor{best}{\textbf{0.594}} 
& \cellcolor{bestbg}\textcolor{best}{\textbf{0.485}} & 
\cellcolor{secondbg}\textcolor{second}{\textbf{454}}
& \cellcolor{secondbg}\textcolor{second}{\textbf{4.825}} & 5.281 
& \cellcolor{bestbg}\textcolor{best}{\textbf{0.546}} 
& \cellcolor{bestbg}\textcolor{best}{\textbf{0.449}} 
& \cellcolor{secondbg}\textcolor{second}{\textbf{643}}
& \cellcolor{secondbg}\textcolor{second}{\textbf{4.861}} & 5.325 & 
\cellcolor{bestbg}\textcolor{best}{\textbf{0.519}} & 
\cellcolor{bestbg}\textcolor{best}{\textbf{0.421}} & 
\cellcolor{secondbg}\textcolor{second}{\textbf{718}} \\

Point4D~\cite{jeon2026point4d}
& 2.750 & 
\cellcolor{secondbg}\textcolor{second}{\textbf{2.688}} & 0.378 & 0.282 & 1176
& 5.211 & 
\cellcolor{secondbg}\textcolor{second}{\textbf{5.229}} & 0.304 & 0.215 & 1669
& 5.334 & 
\cellcolor{secondbg}\textcolor{second}{\textbf{5.306}}
 & 0.266 & 0.178 & 2019 \\
% \midrule

\quad \textbf{+ Ours (Recall4D)}
& \cellcolor{bestbg}\textcolor{best}{\textbf{2.365}} 
& \cellcolor{bestbg}\textcolor{best}{\textbf{2.615}}
& 0.500 & 0.408 
& \cellcolor{bestbg}\textcolor{best}{\textbf{155}}
& \cellcolor{bestbg}\textcolor{best}{\textbf{4.595}} 
& \cellcolor{bestbg}\textcolor{best}{\textbf{5.084}} & 0.466 & 0.380 & 
\cellcolor{bestbg}\textcolor{best}{\textbf{218}} 
& \cellcolor{bestbg}\textcolor{best}{\textbf{4.634}}  & \cellcolor{bestbg}\textcolor{best}{\textbf{5.138}} & 0.448
& 0.362
& \cellcolor{bestbg}\textcolor{best}{\textbf{264}} \\

\bottomrule
\end{tabular}
}
\vspace{-1.3em}
\end{table*}
\noindent\textbf{Implementation Details.} We use Point4D~\cite{jeon2026point4d} as both our backbone and our main baseline. All network weights are kept frozen, and Recall4D involves no training. Long videos are processed in windows of $L=48$ frames with an overlap of 8 frames, \ie, $C_k=\{40k+1,\ldots,40k+48\}$. Consecutive windows are registered by a trimmed Sim(3) fit on the depth of the overlapping frames, excluding pixels inside the foreground masks. Foreground masks are obtained with SAM~3.1~\cite{carion2026sam}. The persistent world state is updated at $m=4$ uniformly spaced keyframes per window. Each foreground instance receives up to $A_o=32$ anchors (24 to 32 in practice), and each query is skinned to its $H=6$ nearest anchors. Unless stated otherwise, all methods are evaluated on up to 8,192 query points per sequence (6.2k on average), sampled among the points visible in the first frame. For fairness, all methods are run on a single NVIDIA H100 (80\,GB) GPU with memory capped at 40\,GB, and runs exceeding this limit are reported as OOM. Runtime is measured as tracker inference, excluding the warm-up run.

\noindent\textbf{Comparison Methods.} We compare Recall4D with two methods designed for long videos, Point4D~\cite{jeon2026point4d} and the concurrent TrackEverything~\cite{jain2026trackeverything} in its sparse ($^\mathsf{s}$) and dense ($^\mathsf{d}$) modes, and with short-horizon methods: the 3D tracker SpatialTrackerV2~\cite{xiao2025spatialtrackerv2} and the feed-forward 4D reconstruction methods V-DPM~\cite{sucar2026v}, 4RC~\cite{luo20264rc}, Any4D~\cite{karhade2026any4d}, TraceAnything~\cite{liu2026trace}, and Track4World~\cite{lu2026track4world}. Following Point4D, SpatialTrackerV2 uses its sliding-window inference, and the feed-forward methods are chained with its selection strategy, which projects each point into every overlapping frame and re-queries it at the frame whose projected depth best matches the predicted depth.

\noindent\textbf{Datasets.} We evaluate Recall4D on two long-sequence benchmarks: PointOdyssey~\cite{zheng2023}, containing videos with an average length of 2,035 frames, and Dynamic Replica~\cite{karaev2023dynamicstereo}, featuring 900-frame test sequences. Both provide dynamic scenes with complex motion, occlusions, and long point trajectories. Diagnosis and ablations use the validation splits, and all comparisons use the test splits.

\noindent\textbf{Metrics.} We report End-Point Error (EPE); the Average Percentage of points within Distance thresholds (APD), averaged over $\delta_{3D} \in \{0.1, 0.3, 0.5, 1.0\}$\,m; Survival Rate (SR)~\cite{jeon2026point4d}, the average fraction of the video a point is tracked before its error first exceeds $\delta_{3D}$, averaged over the same thresholds; EPE$_\text{dyn}$, the EPE of moving queries, whose ground truth departs by more than 0.05 from its first-frame position; and the average inference time per video in seconds (Time). 
Errors are computed in first-frame camera coordinates after one global median scale per video.
\section{Results and Discussion}
\label{sec6}
\subsection{Long-Horizon 4D Tracking}
\label{sec61}
\noindent\textbf{Quantitative comparison.} Tab.~\ref{tab2} compares Recall4D with 3D trackers and feed-forward 4D methods. Recall4D is fastest in all settings, running $7.8$--$9.0\times$ faster than \cite{jeon2026point4d}. At 500 frames, it achieves the best EPE, APD, and SR on both datasets. At 200 frames on PointOdyssey, \cite{xiao2025spatialtrackerv2} achieves lower EPE and \cite{jain2026trackeverything} higher APD and SR, but both are $6$--$20\times$ slower. From 200 to 500 frames, Recall4D's EPE rises from 0.609 to 0.836, versus 0.492 to 1.037 for \cite{xiao2025spatialtrackerv2}, showing better robustness as video length grows.

\begin{figure*}[!t]
    \centering
    \includegraphics[width=\textwidth]{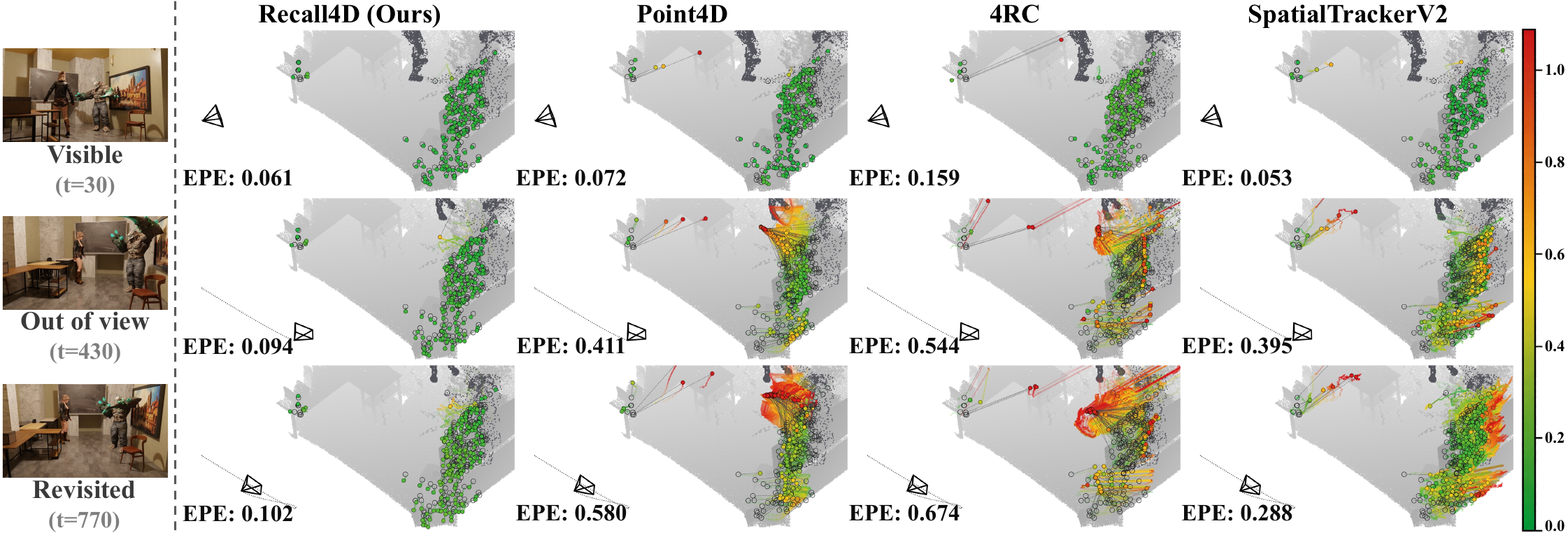}
    \vspace{-2em}
    \caption{\textbf{Qualitative comparison} on PointOdyssey \texttt{scene-d78-3rd} at three moments: visible ($t{=}30$), out of view ($t{=}430$), and revisited ($t{=}770$). Points are colored by error; EPE is shown per panel.}
    \vspace{-1.4em}
    \label{fig5}
\end{figure*}
\noindent\textbf{Long-horizon scalability.}
As Tab.~\ref{tab3} shows, Recall4D lowers the EPE of Point4D by 12--14\%, doubles its SR at 2,000 frames, and runs 7.6$\times$ faster. On TrackEverything, it improves every metric at every length with 14--16\% less runtime, but by smaller margins, since the concurrent TrackEverything also keeps static points in place. This independent convergence supports our diagnosis that forgetting limits long-horizon reconstruction. Absolute errors nevertheless remain high for every method beyond 1,500 frames, a residual shared by all backbones that a training-free plug-in cannot correct; Recall4D removes the forgetting component at a fraction of the cost. We further discuss TrackEverything in detail in Supp.~\ref{secXD1}.

\noindent\textbf{Visualization comparison.}
Fig.~\ref{fig5} shows one PointOdyssey sequence while a region is visible, out of view, and revisited. All methods are accurate while the points are visible. Out of view, Point4D~\cite{jeon2026point4d} and 4RC~\cite{luo20264rc} let static points drift, and re-observation does not repair them. SpatialTrackerV2~\cite{xiao2025spatialtrackerv2} partially recovers upon revisit, whereas Recall4D keeps the points in place throughout.
\begin{table}[t]
\centering
\caption{\textbf{Component ablation} on PointOdyssey val (500 frames). Components are added to chained Point4D. EPE$_\text{dyn}$ is the EPE of moving queries.}
\vspace{-0.8em}
\label{tab4}
\small
\setlength{\tabcolsep}{4pt}
\resizebox{\linewidth}{!}{\begin{tabular}{l ccccc}
\toprule
Setting & EPE$\downarrow$ & APD$\uparrow$ & SR$\uparrow$ & EPE$_\text{dyn}\!\downarrow$ & Time$\downarrow$ \\
\midrule
Baseline (chained Point4D)   & 1.813 & 0.335 & 0.232 & 1.713 & 581.2 \\
\quad + Persistent state     & 1.355 & 0.493 & 0.385 & 1.613 & 288.7 \\
\quad + Sparse anchors       & 1.798 & 0.344 & 0.234 & 1.689 & 347.6 \\
\rowcolor{gray!12}
\quad + Both (\textbf{Recall4D}) & \textbf{1.338} & \textbf{0.499} & \textbf{0.388} & 1.596 & \textbf{66.0} \\
\midrule
\multicolumn{6}{l}{\emph{Variants of Recall4D}} \\
\quad + Per-window rebinding       & 1.348 & 0.495 & 0.388 & 1.604 & 66.1 \\
\quad + Rebinding + retention gate & 1.390 & 0.472 & 0.372 & 1.615 & 237.1 \\
\quad Gate instead of segmenter    & 1.378 & 0.474 & 0.373 & \textbf{1.560} & 410.9 \\
\bottomrule
\end{tabular}}
\vspace{-1.3em}
\end{table}

\subsection{Ablation Studies}
\noindent\textbf{Component ablation.}
Tab.~\ref{tab4} adds each component to chained Point4D~\cite{jeon2026point4d} on PointOdyssey val. The persistent state accounts for nearly all of the accuracy gain, lowering EPE by 25\%. Sparse anchors alone leave accuracy unchanged, including on moving points, while decoding far fewer queries. The two components remove complementary parts of the decoding: each alone saves 40--50\% of the runtime, whereas together they cut it by 8.8$\times$. Neither per-window rebinding nor the retention gate improves on Recall4D, and the gate raises the cost by up to 3.6$\times$. Used instead of the segmenter, the gate attains the lowest EPE$_\text{dyn}$ without any instance masks, but at 6$\times$ the runtime.
\begin{table}[t]
\centering
\caption{\textbf{Decomposition analysis} on 1,000-frame PointOdyssey val videos. Miss and FA are the rates of moving points labeled static and of static points labeled dynamic, per window against ground truth ($\tau{=}0.05$~\cite{jain2026trackeverything}). $^\dagger$Dynamic points decoded densely.}
\vspace{-0.8em}
\label{tab5}
\small
\setlength{\tabcolsep}{3.2pt}
\resizebox{\linewidth}{!}{\begin{tabular}{l cc ccc c}
\toprule
Decomposition & Miss$\downarrow$ & FA$\downarrow$ & EPE$\downarrow$ & EPE$_\text{dyn}\!\downarrow$ & SR$\uparrow$ & Time$\downarrow$ \\
\midrule
\multicolumn{7}{l}{\emph{No decomposition}} \\
\quad All dynamic (Point4D~\cite{jeon2026point4d}) 
& 0   & 100 & 1.569 & 1.172 & 0.193 & 1178 \\
\quad All static (motion discarded)
& 100 & 0   & 0.942 & 1.088 & 0.288 & 103 \\
\midrule
\multicolumn{7}{l}{\emph{First-frame instance masks}} \\
\rowcolor{gray!12}
\quad SAM 3.1~\cite{carion2026sam} (\textbf{default}) 
& 39.5 & 24.1 & \textbf{0.889} & \textbf{1.046} & 0.359 & \textbf{149} \\
\midrule
\multicolumn{7}{l}{\emph{Per-window classification}$^\dagger$} \\
\quad SAM 3.1, first frame        
& 39.5 & 24.1 & 0.916 & 1.079 & 0.355 & 562 \\
\quad SAM 3.1, video propagation       
& 65.5 & \textbf{14.7} & 0.958 & 1.107 & 0.343 & 387 \\
\quad TrackEverything~\cite{jain2026trackeverything} 
& 31.7 & \textbf{14.7} & 0.997 & 1.148 & \textbf{0.372} & 559 \\
\quad Displacement threshold
& 21.3 & 36.2 & 1.084 & 1.137 & 0.324 & 715 \\
\quad Retention gate
& \textbf{7.0} & 58.0 & 1.356 & 1.517 & 0.248 & 471 \\
\midrule
\multicolumn{7}{l}{\emph{Oracle}} \\
\quad GT instance masks
& 9.6 & 50.7 & 0.964 & 1.132 & 0.366 & 280 \\
\quad GT per-window labels
& 0  & 0  & 0.918 & 1.055 & 0.385 & 646 \\
\bottomrule
\end{tabular}}
\vspace{-1.3em}
\end{table}
\begin{figure*}[!t]
    \centering
    \resizebox{\textwidth}{!}{\includegraphics{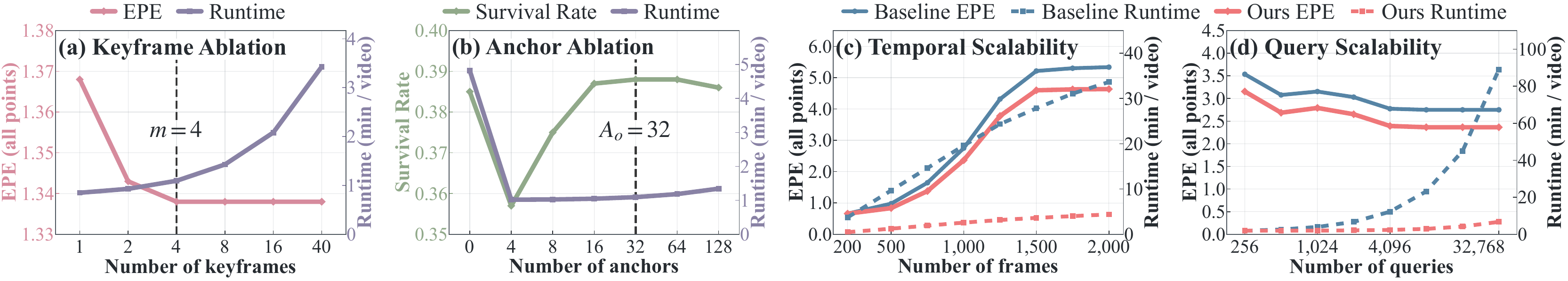}}
    \vspace{-2em}
    % \caption{\textbf{Parameter ablation and scalability.} \textbf{(a, b)} EPE or survival rate and runtime versus the number of keyframes per window $m$ and of anchors per instance $A_o$ on PointOdyssey val (500 frames); $A_o{=}0$ denotes dense decoding without anchors, and dashed lines mark our defaults $m{=}4$ and $A_o{=}32$. \textbf{(c, d)} EPE and runtime of Point4D and Recall4D on PointOdyssey test versus video length (8,192 queries) and number of queries (1,000 frames).}
    \caption{\textbf{Parameter ablation and scalability.} \textbf{(a, b)} Accuracy and runtime versus the number of keyframes per window $m$ and of anchors per instance $A_o$ on PointOdyssey val (500 frames); $A_o{=}0$ is dense decoding, and dashed lines mark our defaults. \textbf{(c, d)} EPE and runtime of Point4D and Recall4D on PointOdyssey test versus video length (8,192 queries) and number of queries (1,000 frames).}
    \vspace{-1.4em}
    \label{fig6}
\end{figure*}

\noindent\textbf{Parameter ablation.}
Fig.~\ref{fig6}\textbf{(a, b)} varies the number of keyframes per window $m$ and of anchors per instance $A_o$. EPE drops from 1.368 at $m{=}1$ to 1.338 at $m{=}4$ and remains unchanged beyond, while runtime grows from 65.7\,s to 205.7\,s at $m{=}40$. With 4 anchors, SR falls to 0.357, below the 0.385 of dense decoding. 
From 16 anchors onward it reaches 0.387 to 0.388, and at $A_o{=}32$, Recall4D runs 4.4$\times$ faster than dense decoding, 66.0\,s against 288.7\,s. We therefore adopt $m{=}4$ and $A_o{=}32$ as our defaults.

\noindent\textbf{Decomposition analysis.}
Tab.~\ref{tab5} replaces Recall4D's first-frame SAM 3.1 decomposition with alternatives. Holding every point discards all motion, yet already lowers EPE from 1.569 to 0.942: memory accounts for most of the gain. Its SR of 0.288, below our 0.359, reflects the frozen moving points, so a decomposition is required. False alarms are the costlier error, since falsely dynamic static points are re-predicted and forget again, raising EPE and lowering SR; the retention gate, with the fewest misses but 58\% false alarms, is the least accurate decomposition. More accurate per-window classification, by TrackEverything~\cite{jain2026trackeverything} or even ground truth, raises SR to at most 0.385 but never lowers EPE. Under the same decomposition, skinning beats dense decoding, 0.889 against 0.916 EPE, at a quarter of the runtime. Recall4D has the lowest EPE and EPE$_\text{dyn}$  overall.
\begin{table}[t]
  \caption{\textbf{Cross-backbone generalization} on PointOdyssey~\cite{zheng2023} (500 frames). Each backbone is evaluated without (w/o) and with (w/) Recall4D. \textcolor{darkgreen}{\textbf{Green}} indicates improvement over the corresponding backbone without Recall4D.}
 \vspace{-0.8em}
  \label{tab6}
  \centering
  \setlength{\tabcolsep}{3pt}
  \resizebox{\linewidth}{!}{%
    \begin{tabular}{lcccc|cccc}
    \toprule
    \multirow{2}{*}{Method}
    & \multicolumn{4}{c}{w/o Recall4D}
    & \multicolumn{4}{c}{w/ Recall4D} \\
    \cmidrule(lr){2-5}
    \cmidrule(lr){6-9}
    & EPE$\downarrow$ & APD$\uparrow$ & SR$\uparrow$ & Time$\downarrow$
    & EPE$\downarrow$ & APD$\uparrow$ & SR$\uparrow$ & Time$\downarrow$ \\
    \midrule
    TraceAnything~\cite{liu2026trace} & 2.572 & 0.137 & 0.072 & 84.3 & \textcolor{darkgreen}{\textbf{1.991}} & \textcolor{darkgreen}{\textbf{0.202}} & \textcolor{darkgreen}{\textbf{0.147}} & \textcolor{darkgreen}{\textbf{75.7}} \\
    Any4D~\cite{karhade2026any4d} & 1.473 & 0.296 & 0.209 & 198.3 & \textcolor{darkgreen}{\textbf{1.303}} & \textcolor{darkgreen}{\textbf{0.324}} & \textcolor{darkgreen}{\textbf{0.254}} & \textcolor{darkgreen}{\textbf{168.0}} \\
    V-DPM~\cite{sucar2026v} & 1.281 & 0.391 & 0.320 & 1377 & \textcolor{darkgreen}{\textbf{1.054}} & \textcolor{darkgreen}{\textbf{0.435}} & \textcolor{darkgreen}{\textbf{0.370}} & \textcolor{darkgreen}{\textbf{1180}} \\
    4RC~\cite{luo20264rc} & 1.626 & 0.347 & 0.283 & 179.8 & \textcolor{darkgreen}{\textbf{1.253}} & \textcolor{darkgreen}{\textbf{0.380}} & \textcolor{darkgreen}{\textbf{0.319}} & \textcolor{darkgreen}{\textbf{144.4}}\\
    Track4World~\cite{lu2026track4world} & 1.689 & 0.303 & 0.183 & 912.5 & \textcolor{darkgreen}{\textbf{1.517}} & \textcolor{darkgreen}{\textbf{0.348}} & \textcolor{darkgreen}{\textbf{0.272}} & \textcolor{darkgreen}{\textbf{733.3}} \\
    \midrule
    \rowcolor{gray!15}
    \textbf{Recall4D (Ours)} & & & &  & \textbf{0.836} & \textbf{0.521} & \textbf{0.440} & \textbf{74.9} \\
    \bottomrule
  \end{tabular}
  }
  \vspace{-1em}
\end{table}

\subsection{Accuracy-Efficiency Analysis}
\label{sec63}
Fig.~\ref{fig6}\textbf{(c, d)} scales the video length and the number of queries. From 200 to 2,000 frames, Recall4D is 7.6--7.8$\times$ faster than Point4D, taking 264\,s instead of 2,019\,s at 2,000 frames, and attains a lower EPE at every length. As the number of queries grows from 256 to 32,768, the runtime of Point4D increases 46.8$\times$, whereas that of Recall4D increases only 3.4$\times$, since added queries are either stored or skinned rather than decoded at every frame. The speed-up thus rises from 7.9$\times$ at 8,192 queries to 13.2$\times$ at 32,768, with a lower EPE at every query budget.

\subsection{Generalization Across Backbones and Tasks}
\noindent\textbf{Cross-backbone generalization.}
To verify that Recall4D's effectiveness is not tied to a specific backbone, we further evaluate its generalization across different feed-forward 4D trackers. Recall4D requires no training and leaves the backbone unchanged, so it applies to other window-chained feed-forward models. Tab.~\ref{tab6} applies it to five further backbones~\cite{
liu2026trace,karhade2026any4d,sucar2026v,luo20264rc,lu2026track4world}, each chained as in Point4D~\cite{jeon2026point4d}. Recall4D improves every metric on every backbone, reducing EPE by 10--23\% and doubling SR on TraceAnything, while also reducing runtime.
\begin{table}[t]
\centering
\caption{\textbf{Cross-task generalization.} Each model is run without and with Recall4D. Task~A: PointOdyssey~\cite{zheng2023}; $\delta_\text{re}$ is $\delta_\text{avg}$ on re-entered points. Task~B: Waymo Open Dataset~\cite{sun2020scalability}. Task~C: KITTI~\cite{geiger2013vision}. \textcolor{darkgreen}{\textbf{Green}} / \textcolor{darkblue}{\textbf{blue}}: better / worse than the backbone.}
\vspace{-0.8em}
\label{tab7}
% \scriptsize
\footnotesize
\setlength{\tabcolsep}{3pt}
\begin{tabular*}{\linewidth}{@{\extracolsep{\fill}}l ccccc@{}}
\toprule
\multicolumn{6}{l}{\emph{Task A: 2D long-range point tracking}} \\
Method & AJ$\uparrow$ & $\delta_\text{avg}\!\uparrow$ & OA$\uparrow$ & $\delta_\text{re}\!\uparrow$ & Time$\downarrow$ \\
\midrule
CoTracker3~\cite{karaev2025cotracker3} & 35.2 & 53.7 & 69.6 & 39.6 & 99 \\
\quad + Recall4D 
& \textcolor{darkgreen}{\textbf{39.3}}  
& \textcolor{darkgreen}{\textbf{58.2}} 
& \textcolor{darkgreen}{\textbf{74.8}} 
& \textcolor{darkgreen}{\textbf{40.3}} 
& \textcolor{darkgreen}{\textbf{72}} \\
\midrule
\multicolumn{6}{l}{\emph{Task B: Long-sequence 3D reconstruction}} \\
Method & Acc$\downarrow$ & Comp$\downarrow$ & CD$\downarrow$ & ATE$\downarrow$ & Time$\downarrow$ \\
\midrule
VGGT-Long~\cite{deng2026vggt} & 1.205 & 2.913 & 2.059 & 2.097 & 237 \\
\quad + Recall4D & \textcolor{darkgreen}{\textbf{1.116}} 
& \textcolor{darkgreen}{\textbf{2.461}} 
& \textcolor{darkgreen}{\textbf{1.789}} 
& \textcolor{darkgreen}{\textbf{1.894}} 
& \textcolor{darkgreen}{\textbf{185}}  \\
\midrule
\multicolumn{6}{l}{\emph{Task C: Video depth estimation}} \\
Method & AbsRel$\downarrow$ & & $\delta_1\!\uparrow$ & & Time$\downarrow$ \\
\midrule
Video Depth Anything~\cite{chen2025video} & 0.092 & & 0.929 & & 63 \\
\quad + Recall4D & \textcolor{darkgreen}{\textbf{0.084}} & & \textcolor{darkgreen}{\textbf{0.957}} & & \textcolor{darkblue}{\textbf{79}} \\
\bottomrule
\end{tabular*}
\vspace{-1em}
\end{table}

\noindent\textbf{Cross-task generalization.} Forgetting is not specific to 4D tracking, so Tab.~\ref{tab7} applies Recall4D to three further tasks. \textbf{\textit{Task A}} is 2D long-range point tracking with CoTracker3~\cite{karaev2025cotracker3}, where static points are read from the persistent world state and projected into the predicted camera. Recall4D raises AJ from 35.2 to 39.3 and $\delta_\text{avg}$ from 53.7 to 58.2 with 27\% less runtime. \textbf{\textit{Task B}} is long-sequence 3D reconstruction with VGGT-Long~\cite{deng2026vggt}, where it lowers the accuracy, completeness, and Chamfer errors by 7\%, 16\%, and 13\% and ATE by 10\%, with 22\% less runtime. \textbf{\textit{Task C}} is video depth estimation with Video Depth Anything~\cite{chen2025video}, where it lowers AbsRel by 9\% and raises $\delta_1$ from 0.929 to 0.957 at 25\% more runtime. Remembering what was observed thus helps beyond 4D tracking.
\section{Conclusion}
Feed-forward 4D trackers degrade on long videos, and we traced this degradation to forgetting rather than inter-window misalignment: ground-truth alignment leaves the error unchanged, whereas simply retaining unseen points reduces it by 26\%. Building on this diagnosis, we introduced Recall4D, a training-free framework that equips a frozen tracker with a persistent world state for static geometry and sparse motion anchors for dynamic objects. Recall4D improves long-horizon accuracy while tracking 2,000 frames 7.6$\times$ faster, and the same principle carries over to other backbones, a concurrent long-video tracker, and three further tasks. Moving objects still depend on the backbone's motion prediction, and extending memory to them is a natural next step. We hope these findings encourage long-video 4D reconstruction to treat memory, and not only alignment, as a core design problem.

\clearpage  
\section*{LLM Usage Statement}
ChatGPT (OpenAI) and Claude (Anthropic) were used to assist with English language editing, grammatical refinement, and improving the clarity, coherence, and organization of the manuscript. Generative AI tools were not used for synthetic data generation, research methodology development, experimental design, or the analysis and interpretation of experimental results. All AI-assisted revisions were carefully reviewed to ensure technical accuracy and consistency with the intended scientific meaning. The authors retain full responsibility for the scientific ideas, methodological contributions, experimental findings, and conclusions presented in this paper.
{
    \small
    \bibliographystyle{ieeenat_fullname}
    \bibliography{ref}
}
\clearpage
\maketitlesupplementary
\setcounter{section}{0}
\renewcommand{\thesection}{\Alph{section}}
\renewcommand{\theHsection}{supp.\Alph{section}}

\section*{Overview}
The supplementary material is structured as follows:
\begin{itemize}
\item Section~\ref{secXA}. Additional Details on Error Analysis.
\item Section~\ref{secXB}. Implementation and Evaluation Details.
\item Section~\ref{secXC}. Additional Experiments.
\item Section~\ref{secXD}. Discussion and Limitations.
\item Section~\ref{secXE}. Extended Visualizations.
\end{itemize}

\begin{table*}[t]
\centering
\setlength{\tabcolsep}{4pt}
\caption{\textbf{Error analysis under the standard protocol.} The experiment of Tab.~\ref{tab1} with all GT tracks retained. Only the dynamic out-of-frame class changes: it is dominated by the scene-exiting GT tracks of \texttt{ani14\_new\_f} and \texttt{r1\_new\_f}, which all methods inherit alike.}
\vspace{-0.8em}
\label{tab8}
\resizebox{\linewidth}{!}{\begin{tabular}{c l ccc ccc ccc}
\toprule
 &  & \multicolumn{3}{c}{All points} & \multicolumn{3}{c}{Static points} & \multicolumn{3}{c}{Dynamic points} \\
\cmidrule(lr){3-5}\cmidrule(lr){6-8}\cmidrule(lr){9-11}
 &  & EPE$\downarrow$ & APD$\uparrow$ & SR$\uparrow$ & visible & occluded & out of frame & visible & occluded & out of frame \\
\midrule
 & Share of (point, frame) pairs & 100\% & -- & -- & 8.3\% & 2.4\% & 17.0\% & 40.1\% & 17.8\% & 14.5\% \\
\midrule
 & Chained baseline & 3.831 & 0.280 & 0.186 & 1.246 & 1.162 & 1.798 & 1.196 & 1.279 & 14.607 \\

(a) & \quad + GT-pose alignment (no registration drift) & 3.832 {\scriptsize(0\%)} & 0.287 {\scriptsize(+2\%)} & 0.192 {\scriptsize(+3\%)} & 1.252 {\scriptsize(0\%)} & 1.189 {\scriptsize(+2\%)} & 1.878 {\scriptsize(+4\%)} & 1.168 {\scriptsize($-$2\%)} & 1.244 {\scriptsize($-$3\%)} & 14.626 {\scriptsize(0\%)} \\

(b) & \quad + hold unseen points (predicted view) & 3.430 {\scriptsize($-$10\%)} & 0.383 {\scriptsize(+37\%)} & 0.253 {\scriptsize(+36\%)} & 1.028 {\scriptsize($-$17\%)} & 1.024 {\scriptsize($-$12\%)} & 1.115 {\scriptsize($-$38\%)} & 1.115 {\scriptsize($-$7\%)} & 1.195 {\scriptsize($-$7\%)} & 14.380 {\scriptsize($-$2\%)} \\

(d) & \quad + hold unseen points (GT visibility) & 3.306 {\scriptsize($-$14\%)} & 0.435 {\scriptsize(+55\%)} & 0.275 {\scriptsize(+48\%)} & 0.971 {\scriptsize($-$22\%)} & 0.897 {\scriptsize($-$23\%)} & 0.766 {\scriptsize($-$57\%)} & 1.063 {\scriptsize($-$11\%)} & 1.164 {\scriptsize($-$9\%)} & 14.333 {\scriptsize($-$2\%)} \\
\midrule
\rowcolor{gray!12}
 & \textbf{Recall4D (Ours)} & \textbf{3.157} {\scriptsize($-$18\%)} & \textbf{0.469} {\scriptsize(+67\%)} & 0.353 {\scriptsize(+90\%)} & \textbf{0.435} {\scriptsize($-$65\%)} & \textbf{0.498} {\scriptsize($-$57\%)} & 0.434 {\scriptsize($-$76\%)} & \textbf{0.954} {\scriptsize($-$20\%)} & \textbf{1.066} {\scriptsize($-$17\%)} & \textbf{14.303} {\scriptsize($-$2\%)} \\
 & Recall4D with GT masks and GT visibility & 3.231 {\scriptsize($-$16\%)} & 0.454 {\scriptsize(+62\%)} & 0.361 {\scriptsize(+94\%)} & 0.460 {\scriptsize($-$63\%)} & 0.512 {\scriptsize($-$56\%)} & 0.377 {\scriptsize($-$79\%)} & 1.014 {\scriptsize($-$15\%)} & 1.107 {\scriptsize($-$13\%)} & 14.466 {\scriptsize($-$1\%)} \\
\bottomrule
\end{tabular}}
\vspace{-0.5em}
\end{table*}
\begin{table*}[t]
\centering
\setlength{\tabcolsep}{8pt}
\caption{\textbf{Recall4D with oracle inputs} on PointOdyssey val (1,000 frames), under the protocol of Tab.~\ref{tab1}. Brackets give the change relative to Recall4D. Replacing the SAM~3.1 masks by GT masks of moving objects, or the projected visibility by the GT one, does not lower the overall EPE: the remaining error does not come from the perception inputs.}
\vspace{-0.8em}
\label{tab9}
\resizebox{\linewidth}{!}{\begin{tabular}{l ccc ccc ccc}
\toprule
 & \multicolumn{3}{c}{All points} & \multicolumn{3}{c}{Static points} & \multicolumn{3}{c}{Dynamic points} \\
\cmidrule(lr){2-4}\cmidrule(lr){5-7}\cmidrule(lr){8-10}
 & EPE$\downarrow$ & APD$\uparrow$ & SR$\uparrow$ & visible & occluded & out of frame & visible & occluded & out of frame \\
\midrule
\rowcolor{gray!12}
\textbf{Recall4D (Ours)} & \textbf{0.889} & \textbf{0.471} & 0.359 & \textbf{0.434} & \textbf{0.497} & 0.433 & \textbf{0.948} & \textbf{1.064} & \textbf{1.534} \\
\quad with GT masks & 0.964 {\scriptsize(+8\%)} & 0.454 {\scriptsize($-$4\%)} & 0.366 {\scriptsize(+2\%)} & 0.455 {\scriptsize(+5\%)} & 0.508 {\scriptsize(+2\%)} & 0.439 {\scriptsize(+1\%)} & 1.010 {\scriptsize(+7\%)} & 1.104 {\scriptsize(+4\%)} & 1.692 {\scriptsize(+10\%)} \\
\quad with GT visibility & 0.897 {\scriptsize(+1\%)} & 0.470 {\scriptsize(0\%)} & 0.361 {\scriptsize(+1\%)} & 0.444 {\scriptsize(+2\%)} & 0.504 {\scriptsize(+2\%)} & 0.367 {\scriptsize($-$15\%)} & 0.951 {\scriptsize(0\%)} & 1.075 {\scriptsize(+1\%)} & 1.565 {\scriptsize(+2\%)} \\
\quad with GT masks and GT visibility & 0.965 {\scriptsize(+9\%)} & 0.455 {\scriptsize($-$3\%)} & 0.368 {\scriptsize(+3\%)} & 0.460 {\scriptsize(+6\%)} & 0.513 {\scriptsize(+3\%)} & 0.377 {\scriptsize($-$13\%)} & 1.009 {\scriptsize(+7\%)} & 1.107 {\scriptsize(+4\%)} & 1.705 {\scriptsize(+11\%)} \\
\bottomrule
\end{tabular}}
\vspace{-1em}
\end{table*}
\section{Additional Details on Error Analysis}
\label{secXA}

This section details the analysis of Sec.~\ref{sec3}: the point classes behind Tab.~\ref{tab1} and Fig.~\ref{fig3}, the treatment of scene-exiting ground-truth tracks, and how each intervention is implemented. Every intervention is applied to Point4D's own output, with the released weights unchanged.

\noindent\textbf{Point classes.} The analysis uses the queries and metrics of Sec.~\ref{sec5}. Each (query, frame) pair is assigned one of three ground-truth states: \emph{visible}, \emph{occluded}, meaning in front of the camera and inside the image but hidden, or \emph{out of frame}. Together with the static or dynamic label of the query, this gives the six classes of Tab.~\ref{tab1}. The EPE of a class is computed per video over the pairs of that class, then averaged over the videos with at least 100 such pairs. The EPE over all points is averaged over videos in the same way, so it is not the share-weighted mean of the class values.

\noindent\textbf{Scene-exiting ground-truth tracks.} Some PointOdyssey tracks leave the scene: their ground-truth positions jump to hundreds or thousands of meters from the rest of the scene. Let $r$ be the median distance of a video's valid ground-truth positions to their median point. A track is marked as scene-exiting if a valid position lies farther than $25r$ from that point, or if it moves by more than $5r$ between two consecutive valid frames. This removes 145 of 5,979 tracks in \texttt{ani14\_new\_f} and 462 of 6,642 tracks in \texttt{r1\_new\_f}, all on moving objects, and none elsewhere. The same tracks are removed for every method. They do not affect any conclusion but dominate the dynamic out-of-frame class: as Tab.~\ref{tab8} shows, under the standard protocol its EPE exceeds 14 for every method, and the overall EPE of Point4D rises from 1.569 to 3.831.

\noindent\textbf{GT-pose alignment.} Chained Point4D expresses each window $k$ in the predicted camera frame of its first frame $t_k$, in the window's own predicted scale. At each handover it estimates a Sim(3) from the predicted depth of the overlapping frames, which both maps the carried queries into the next window and is accumulated into the world-to-window transform $G_k$ of Sec.~\ref{sec41}. The oracle replaces this estimate by the one implied by the ground-truth camera poses. With $W_t$ the ground-truth world-to-camera transform and $s_k$ the scale of window $k$, measured as the median ratio between predicted and ground-truth camera distances of the tracks visible at $t_k$, a point $\mathbf{x}$ in window $k$ maps to
\begin{equation}
\begin{aligned}
  &\mathbf{x}' = \frac{s_{k+1}}{s_k}\,\mathbf{R}_M\,\mathbf{x}
  + s_{k+1}\,\mathbf{t}_M, \\
  &[\mathbf{R}_M \,|\, \mathbf{t}_M]
  = W_{t_{k+1}} W_{t_k}^{-1}.
\end{aligned}
\end{equation}
in window $k{+}1$. The chain then carries no registration error, while every prediction inside a window remains Point4D's own. Over the 24 handovers of a 1,000-frame video, Point4D's estimates differ from the oracle ones by $1.2^\circ$ in rotation, $3.4\%$ in scale, and 0.10\,m in translation per handover, on average.

\noindent\textbf{Holding unseen points.} Given any tracker's output, a point that has been unseen for more than 40 consecutive frames, the stride between windows, is output at its last seen position until it is seen again. Within a window, the decoder still observes the point in neighboring frames, so the hold starts only after the point has been absent for a full stride. We use two notions of \emph{seen}:
\begin{itemize}
  \item \emph{Predicted view}: the output projects in front of Point4D's predicted camera and inside the image. Since the held output is frozen, the point is released once the \emph{held} position projects back into the view. This variant uses no ground truth, but cannot detect occlusions.
  \item \emph{GT visibility}: the ground-truth visibility flag, which also covers occlusions. This isolates the effect of the hold from that of the visibility decision.
\end{itemize}
The hold is applied to all points, static and dynamic alike, without any segmentation. Row~(c) of Tab.~\ref{tab1} combines the GT-pose alignment with the predicted-view hold. The attempts to repair alignment mentioned in Sec.~\ref{sec3} are described in Sec.~\ref{secXD2}.

\noindent\textbf{Curves of Fig.~\ref{fig3}.} Both panels use the same retained points as Tab.~\ref{tab1}. In (a), the EPE of all points is computed per video in each 40-frame window and averaged over the 15 videos; all of them have 1,000 frames, so every window averages the same videos. The bars give the fraction of retained pairs that are unseen in that window. In (b), each pair of a static point is binned by the number of frames since the point was last visible according to the ground truth (0 when visible); a bin enters a video's mean with at least 50 pairs, and the bars give the share of static pairs per bin.

\noindent\textbf{Recall4D with oracle inputs.} Tab.~\ref{tab9} replaces the two perception inputs of Recall4D by ground truth, one at a time and jointly. \emph{GT masks} use the PointOdyssey instance masks, restricted in each window to objects that move within it: an object counts as moving when at least 10\% of the ground-truth tracks projecting onto it move by more than 0.05\,m in that window, and every other object is treated as background, as SAM~3.1 does with its text prompts. This is the setting of the GT instance masks row of Tab.~\ref{tab5}. \emph{GT visibility} replaces the projection test of Eq.~(4): at each keyframe, the persistent state of a point is updated only if the point is visible according to the ground truth. In the first window every point is still updated, as in the method. Neither oracle lowers the overall EPE, and SR changes by less than 0.01; the remaining error does not come from the perception inputs.
\section{Implementation and Evaluation Details}
\label{secXB}
\noindent\textbf{Fig.~\ref{fig1}.} The video in \textbf{(a)} was recorded by the authors with a handheld iPhone and is not part of any dataset. In \textbf{(b)}, we report the survival rate at 0.1\,m, the strictest of our four thresholds, whereas the SR in our tables averages over all four. The baseline in \textbf{(b)} and \textbf{(c)} is chained Point4D.

\noindent\textbf{Fig.~\ref{fig2}.} The video was recorded by the authors with a handheld iPhone. We track 64 points on a parked car and draw three of them. The car is static, so the reference for each point is its first-frame position from the same tracker. The raw error is the mean distance to this reference over the 64 points. For the Sim(3) curve, we fit one transform per frame with the Umeyama algorithm and report the mean residual. This setting favors the misalignment hypothesis, since the transform may change at every frame. Errors are divided by the car size, the mean distance of the first-frame points to their centroid. The baseline error stays high even after the per-frame fit, so a similarity transform cannot explain it. The error of Recall4D stays near zero. The left panel is a schematic of misalignment: the points drift together and keep their relative positions. The middle panel shows the baseline tracks, which scatter independently. The right panel shows Recall4D, which keeps the points in place.
\begin{figure*}[!t]
    \centering
    \resizebox{\textwidth}{!}{\includegraphics{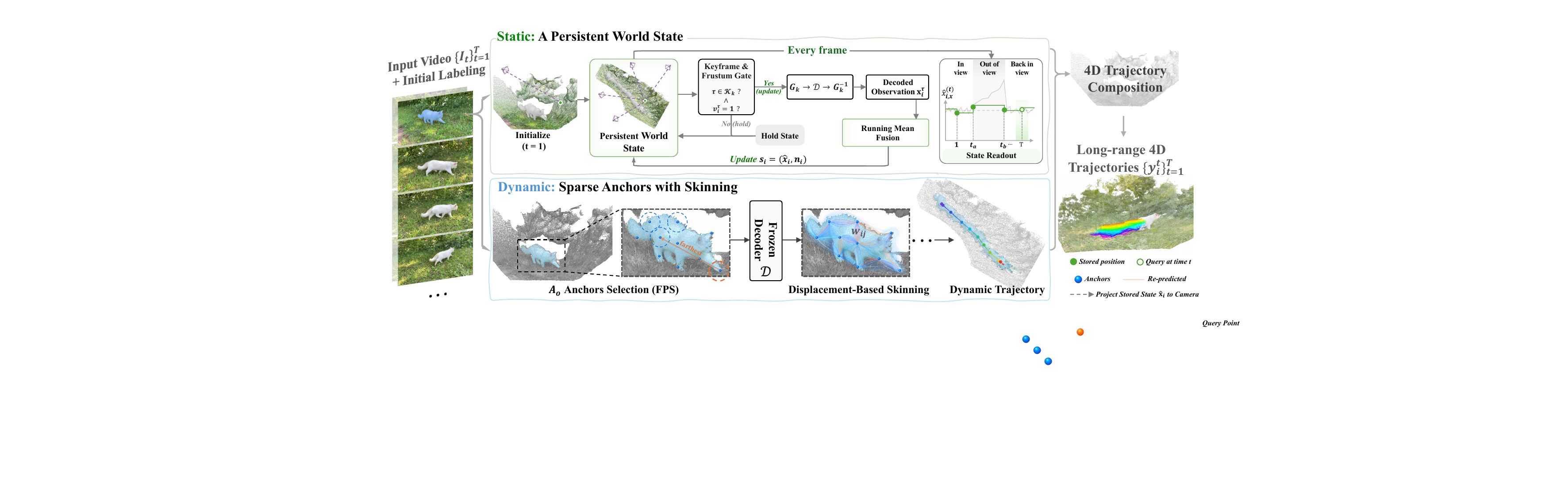}}
    \vspace{-2em}
    \caption{\textbf{Detailed pipeline of the persistent world state.} At each keyframe, a stored point that lies inside the camera frustum is decoded and fused into the state by a running mean; otherwise the state is held. The output at every frame is read from the state.}
    \vspace{-1em}
    \label{fig8}
\end{figure*}
\begin{figure}[t]
\setlength{\intextsep}{0pt}
\begin{algorithm}[H]
\caption{\textbf{Recall4D inference}}
\label{alg1}
\small
\begin{algorithmic}[1]
\Require video $\{I_t\}_{t=1}^{T}$, queries $\{\mathbf{u}_i\}_{i=1}^{N}$, frozen encoder $\mathcal{E}$ and decoder $\mathcal{D}$, first-frame instance masks $M_1$
\Ensure trajectories $\{\mathbf{y}_i^t\}$
\State $\mathcal{B} \gets \{i : M_1(\mathbf{u}_i)=0\}$ \Comment{background queries}
\State lift each $i\in\mathcal{B}$ to $\hat{\mathbf{x}}_i$ with predicted depth;\; $n_i \gets 0$
\For{each foreground instance $o$}
  \State $\{\bar{\mathbf{a}}_j\} \gets \mathrm{FPS}(X_o, A_o)$ \Comment{anchors}
  \State compute $\mathcal{N}(i)$ and $w_{ij}$ for its queries \Comment{once}
\EndFor
\For{each window $C_k$}
  \State encode $\mathcal{E}(C_k)$; estimate cameras and $G_k$
  \For{each keyframe $\tau \in \mathcal{K}_k$} \Comment{static: update}
    \For{each $i \in \mathcal{B}$ with $v_i^\tau = 1$} \Comment{others are held}
      \State decode $\mathbf{x}_i^\tau$ from the stored $\hat{\mathbf{x}}_i$
      \State $\hat{\mathbf{x}}_i \gets (n_i\hat{\mathbf{x}}_i+\mathbf{x}_i^\tau)/(n_i+1)$;\; $n_i \gets n_i+1$
    \EndFor
  \EndFor
  \For{each frame $t \in C_k$}
    \State $\mathbf{y}_i^t \gets \hat{\mathbf{x}}_i$ for $i\in\mathcal{B}$ \Comment{static: read out}
    \State decode anchors $\mathbf{a}_j^t$ \Comment{dynamic: track}
    \State $\mathbf{y}_i^t \gets \bar{\mathbf{x}}_i^{(k)} + \sum_{j} w_{ij}(\mathbf{a}_j^t - \bar{\mathbf{a}}_j^{(k)})$ for $i\notin\mathcal{B}$ \Comment{skin}
  \EndFor
  \State update rest positions $\bar{\mathbf{x}}_i^{(k+1)}, \bar{\mathbf{a}}_j^{(k+1)}$
\EndFor
\end{algorithmic}
\end{algorithm}
\vspace{-1.2em}
\end{figure}

\noindent\textbf{Tab.~\ref{tab1} \& Fig.~\ref{fig3}.} Their implementation is detailed in Sec.~\ref{secXA}.

\noindent\textbf{Fig.~\ref{fig4}.} The cat video used for illustration is an in-the-wild clip from the internet, about 30 seconds long. Fig.~\ref{fig8} details the persistent world state, showing the complete pipeline for static background points, and Alg.~\ref{alg1} summarizes the full inference procedure of Recall4D.

\noindent\textbf{Tab.~\ref{tab2}.} The 200- and 500-frame settings use the first 200 and 500 frames of each test sequence. The Recall4D row is Point4D with Recall4D. TrackEverything runs in its dense mode. In all our experiments, its input pointmaps come from VGGT-$\Omega$~\cite{wang2026vggtomega}.

\noindent\textbf{Tab.~\ref{tab3}.} At these lengths we use the sparse mode of TrackEverything, which fits within our memory cap. TrackEverything already classifies each point as static or dynamic in every window. We keep this classification and apply Recall4D on top of it: points classified as static are written to the persistent world state and read out from it, and dynamic points are tracked through sparse anchors with skinning. Test sequences shorter than the target length are evaluated on all their frames.

\noindent\textbf{Fig.~\ref{fig5}.} Each panel is a top-down view of the reconstructed scene, with the current camera drawn as a frustum. We draw the static queries in the region that leaves the view. Filled dots are the predictions, colored by their error at that frame, and open circles are the ground-truth positions, connected to their predictions by lines. The color scale saturates at 1.0. The EPE in each panel is the mean error over the points.

\noindent\textbf{Tab.~\ref{tab4}.} All rows use the first 500 frames of the validation videos, under the standard protocol with all tracks. \emph{Persistent state} alone stores the static points and decodes every dynamic query; \emph{sparse anchors} alone skins the dynamic queries and re-predicts the static points. \emph{Per-window rebinding} recomputes the anchor associations and skinning weights at every window. The \emph{retention gate} decides which points enter the state: a small MLP, trained on PointOdyssey training sequences, classifies each point once from nine depth-normalized features of its predicted trajectory in the first window. With SAM 3.1, a point enters the state only if it lies in the background and passes the gate; \emph{gate instead of segmenter} uses the gate alone.
\begin{table*}[t]
\centering
\setlength{\tabcolsep}{5pt}
\caption{\textbf{Long-horizon comparison without scene-exiting tracks.} Tab.~\ref{tab3} re-evaluated on PointOdyssey test with the track filter of Sec.~\ref{secXA}. Predictions are unchanged. Brackets give the change in EPE relative to the backbone.}
\vspace{-0.8em}
\label{tab10}
\resizebox{\linewidth}{!}{\begin{tabular}{l cccc cccc cccc}
\toprule
\multirow{2}{*}{Method}
 & \multicolumn{4}{c}{1,000 frames} & \multicolumn{4}{c}{1,500 frames} & \multicolumn{4}{c}{2,000 frames} \\
\cmidrule(lr){2-5}\cmidrule(lr){6-9}\cmidrule(lr){10-13}
 & EPE$\downarrow$ & EPE$_\text{dyn}\!\downarrow$ & APD$\uparrow$ & SR$\uparrow$
 & EPE$\downarrow$ & EPE$_\text{dyn}\!\downarrow$ & APD$\uparrow$ & SR$\uparrow$
 & EPE$\downarrow$ & EPE$_\text{dyn}\!\downarrow$ & APD$\uparrow$ & SR$\uparrow$ \\
\midrule
Tracks removed & \multicolumn{4}{c}{2{,}940 of 80{,}799} & \multicolumn{4}{c}{2{,}934 of 80{,}799} & \multicolumn{4}{c}{2{,}934 of 80{,}799} \\
\midrule
TrackEverything$^s$~\cite{jain2026trackeverything} & 1.013 & 1.154 & 0.591 & 0.478 & 1.126 & 1.301 & 0.534 & 0.435 & 1.166 & 1.358 & 0.501 & 0.398 \\
\rowcolor{gray!8}
\quad \textbf{+ Recall4D (Ours)} & 1.006 {\scriptsize(--0.7\%)} & 1.152 & 0.597 & 0.488 & 1.114 {\scriptsize(--1.1\%)} & 1.299 & 0.549 & 0.453 & 1.150 {\scriptsize(--1.4\%)} & 1.354 & 0.522 & 0.425 \\
Point4D~\cite{jeon2026point4d} & 1.165 & 0.969 & 0.378 & 0.282 & 1.481 & 1.137 & 0.304 & 0.215 & 1.598 & 1.194 & 0.266 & 0.178 \\
\rowcolor{gray!8}
\quad + \textbf{Recall4D (Ours)} & 0.805 {\scriptsize(--31\%)} & 0.990 & 0.502 & 0.410 & 0.886 {\scriptsize(--40\%)} & 1.119 & 0.468 & 0.382 & 0.924 {\scriptsize(--42\%)} & 1.168 & 0.451 & 0.364 \\
\bottomrule
\end{tabular}}
\vspace{-0.5em}
\end{table*}
\begin{table*}[t]
\centering
\setlength{\tabcolsep}{4pt}
\caption{\textbf{The all-static baseline across video lengths and datasets.} All static holds every query, as in Tab.~\ref{tab5}. SR$_\text{dyn}$ is the SR of dynamic queries, and the first row gives their share.}
\vspace{-0.8em}
\label{tab11}
\resizebox{\linewidth}{!}{\begin{tabular}{l ccc ccc ccc | ccc ccc | ccc}
\toprule
\multirow{2}{*}{Method}
 & \multicolumn{3}{c}{PointOdyssey (200)} & \multicolumn{3}{c}{PointOdyssey (500)} & \multicolumn{3}{c}{PointOdyssey (1,000)}
 & \multicolumn{3}{c}{Dynamic Replica (200)} & \multicolumn{3}{c}{Dynamic Replica (500)}
 & \multicolumn{3}{c}{ADT (300)} \\
\cmidrule(lr){2-4}\cmidrule(lr){5-7}\cmidrule(lr){8-10}\cmidrule(lr){11-13}\cmidrule(lr){14-16}\cmidrule(lr){17-19}
 & EPE$\downarrow$ & SR$\uparrow$ & SR$_\text{dyn}\!\uparrow$ & EPE$\downarrow$ & SR$\uparrow$ & SR$_\text{dyn}\!\uparrow$ & EPE$\downarrow$ & SR$\uparrow$ & SR$_\text{dyn}\!\uparrow$
 & EPE$\downarrow$ & SR$\uparrow$ & SR$_\text{dyn}\!\uparrow$ & EPE$\downarrow$ & SR$\uparrow$ & SR$_\text{dyn}\!\uparrow$
 & EPE$\downarrow$ & SR$\uparrow$ & SR$_\text{dyn}\!\uparrow$ \\
\midrule
Dynamic queries & \multicolumn{3}{c}{59.2\%} & \multicolumn{3}{c}{59.7\%} & \multicolumn{3}{c}{59.7\%} & \multicolumn{3}{c}{32.3\%} & \multicolumn{3}{c}{34.1\%} & \multicolumn{3}{c}{27.7\%} \\
Chained baseline 
& 0.661 & 0.489 & 0.477 
& 0.979 & 0.370 & 0.377 
& 2.750 & 0.282 & 0.245 
& 0.255 & 0.710 & 0.653 
& 0.307 & 0.646 & 0.583 
& 0.870 & 0.412 & 0.381 \\
All static       
& 0.653 & 0.483 & 0.438 
& 0.838 & 0.428 & 0.351 
& 2.370 & 0.378 & 0.229 
& 0.250 & 0.703 & 0.557 
& 0.270 & 0.660 & 0.464 
& 0.638 & 0.464 & 0.351 \\
\rowcolor{gray!12}
\textbf{Recall4D (Ours)} 
& 0.609 & 0.528 & 0.471 
& 0.836 & 0.440 & 0.379 
& 2.365 & 0.408 & 0.287 
& 0.244 & 0.725 & 0.639 
& 0.262 & 0.690 & 0.572 
& 0.629 & 0.474 & 0.386 \\
\bottomrule
\end{tabular}}
\vspace{-1em}
\end{table*}
\begin{table*}[t]
  \caption{\textbf{Quantitative comparison} on TAPVid-3D ADT minival split. Each backbone is evaluated without and with Recall4D. $\Delta$ gives the change brought by Recall4D, in \textcolor{darkgreen}{\textbf{green}} when it is an improvement.}
  \vspace{-0.8em}
  \label{tab12}
  \centering\footnotesize
  \setlength{\tabcolsep}{3pt}
  \newcommand{\ours}{\rowcolor{gray!8}\quad \textbf{+ Recall4D (Ours)}}
  \begin{tabularx}{\linewidth}{l *{8}{>{\centering\arraybackslash}X}}
    \toprule
    Method & EPE$\downarrow$ & $\Delta$EPE & APD$\uparrow$ & $\Delta$APD & SR$\uparrow$ & $\Delta$SR & Time\,(s)$\downarrow$ & $\Delta$Time \\
    \midrule

    SpatialTrackerV2~\cite{xiao2025spatialtrackerv2}
    & 1.304 & -- & 0.624 & -- & 0.546 & -- & 220.9 & -- \\

    TraceAnything~\cite{liu2026trace}
    & 0.851 & -- & 0.523 & -- & 0.476 & -- & 78.2 & -- \\
    \ours
    & 0.518
    & \textcolor{darkgreen}{\textbf{$-$0.333}}
    & 0.682
    & \textcolor{darkgreen}{\textbf{$+$0.159}}
    & 0.660
    & \textcolor{darkgreen}{\textbf{$+$0.184}}
    & 66.6
    & \textcolor{darkgreen}{\textbf{$-$11.6}} \\

    Any4D~\cite{karhade2026any4d}
    & 0.817 & -- & 0.408 & -- & 0.364 & -- & 199.0 & -- \\
    \ours
    & 0.611
    & \textcolor{darkgreen}{\textbf{$-$0.206}}
    & 0.512
    & \textcolor{darkgreen}{\textbf{$+$0.104}}
    & 0.495
    & \textcolor{darkgreen}{\textbf{$+$0.131}}
    & 54.5
    & \textcolor{darkgreen}{\textbf{$-$144.5}} \\

    V-DPM~\cite{sucar2026v}
    & 0.714 & -- & 0.534 & -- & 0.488 & -- & 2325 & -- \\
    \ours
    & 0.417
    & \textcolor{darkgreen}{\textbf{$-$0.297}}
    & 0.731
    & \textcolor{darkgreen}{\textbf{$+$0.197}}
    & 0.714
    & \textcolor{darkgreen}{\textbf{$+$0.226}}
    & 2106
    & \textcolor{darkgreen}{\textbf{$-$219}} \\

    4RC~\cite{luo20264rc}
    & 0.724 & -- & 0.545 & -- & 0.492 & -- & 155.8 & -- \\
    \ours
    & 0.398
    & \textcolor{darkgreen}{\textbf{$-$0.326}}
    & 0.721
    & \textcolor{darkgreen}{\textbf{$+$0.176}}
    & 0.705
    & \textcolor{darkgreen}{\textbf{$+$0.213}}
    & 75.8
    & \textcolor{darkgreen}{\textbf{$-$80.0}} \\

    Track4World~\cite{lu2026track4world}
    & 0.909 & -- & 0.383 & -- & 0.320 & -- & 478.8 & -- \\
    \ours
    & 0.635
    & \textcolor{darkgreen}{\textbf{$-$0.274}}
    & 0.497
    & \textcolor{darkgreen}{\textbf{$+$0.114}}
    & 0.483
    & \textcolor{darkgreen}{\textbf{$+$0.163}}
    & 124.0
    & \textcolor{darkgreen}{\textbf{$-$354.8}} \\

    TrackEverything$^\mathsf{s}$~\cite{jain2026trackeverything}
    & 0.273 & -- & 0.768 & -- & 0.740 & -- & 235.3 & -- \\
    \ours
    & 0.270
    & \textcolor{darkgreen}{\textbf{$-$0.003}}
    & 0.769
    & \textcolor{darkgreen}{\textbf{$+$0.001}}
    & 0.741
    & \textcolor{darkgreen}{\textbf{$+$0.001}}
    & 221.9
    & \textcolor{darkgreen}{\textbf{$-$13.4}} \\

    Point4D~\cite{jeon2026point4d}
    & 0.870 & -- & 0.451 & -- & 0.412 & -- & 42.7 & -- \\
    \ours
    & 0.629
    & \textcolor{darkgreen}{\textbf{$-$0.241}}
    & 0.493
    & \textcolor{darkgreen}{\textbf{$+$0.042}}
    & 0.474
    & \textcolor{darkgreen}{\textbf{$+$0.062}}
    & 26.8
    & \textcolor{darkgreen}{\textbf{$-$15.9}} \\

    \bottomrule
  \end{tabularx}
  \vspace{-1em}
\end{table*}

\noindent\textbf{Tab.~\ref{tab5}.} All rows use the 1,000-frame validation videos with scene-exiting tracks excluded, as in Tab.~\ref{tab1}. A point is dynamic in a window if the bounding box of its ground-truth positions there has a diagonal above 0.05\,m; Miss and FA are measured against these labels. Runtimes exclude the segmentation or classification itself.

\textit{All dynamic} is chained Point4D, and \textit{All static} holds every point at its stored position. The rows under per-window classification relabel the points in every window and decode all dynamic points, without anchors. \textit{SAM 3.1, first frame} keeps the default labels, so it isolates the effect of skinning. \textit{Video propagation} propagates the SAM 3.1 masks through the video and, at the start of each window, labels a point as dynamic if its current estimate projects into an instance mask; a point outside the image keeps its label. \textit{TrackEverything} uses its predicted labels. \textit{Displacement threshold} labels a point as dynamic if the bounding-box diagonal of its predicted positions in the previous window exceeds 2\% of the median query depth, a threshold chosen on the Dynamic Replica validation set. \textit{Retention gate} uses the gate of Tab.~\ref{tab4} and, unlike the other rows of this group, labels the points once, from the first window. \textit{GT instance masks} are described in Sec.~\ref{secXA}, and \textit{GT per-window labels} use the ground-truth window labels defined above.

\noindent\textbf{Tab.~\ref{tab6} \& Fig.~\ref{fig6}.} Their implementation is detailed in Sec.~\ref{sec6}.

\noindent\textbf{Tab.~\ref{tab7}.} Each task attaches the persistent world state of Sec.~4.2 to an existing model, with the adaptations below.

\textit{Task A.} CoTracker3 tracks in 2D, so the state is kept in 3D and projected back into the image. Points outside the SAM 3.1 instances of the first frame are static, and are lifted to 3D and stored as in Sec.~4.2. In every frame, the camera is registered to the state by PnP with RANSAC on the 2D positions from the tracker, with a 4-pixel threshold; if too few inliers are found, the camera predicted by Point4D is used. The state supplies the position of a static point only while the tracker does not see it. A point that the registration repeatedly rejects as an outlier is treated as a moving point missed by the masks and returned to the tracker. Visibility always comes from the tracker.

\textit{Task B.} VGGT-Long~\cite{deng2026vggt} predicts a pointmap per chunk and stacks the Sim(3)-aligned chunks, so a region is predicted again whenever a new chunk sees it. We keep a world-space state for static regions instead: observations are fused by a running mean at keyframes, and a region that leaves the view keeps its stored geometry. Vehicles and pedestrians are removed with SAM 3.1 masks, and no anchors are needed since the benchmark evaluates the static scene.

\textit{Task C.} Video Depth Anything predicts depth independently in each window, so the depth of a static region changes between windows and upon revisits. We back-project static pixels into world coordinates with estimated camera poses, fuse them into the state by a running mean at keyframes, and read their depth back by projection. Dynamic pixels, given by SAM 3.1 masks, keep the original prediction, so no anchors are needed.
\section{Additional Experiments}
\label{secXC}
This section reports three experiments: long-horizon results without scene-exiting tracks, a closer look at the all-static baseline, and an evaluation on real-world videos.

\subsection{Results without Scene-Exiting Tracks}
\label{secXC1}
In Tab.~\ref{tab3}, Recall4D lowers the EPE of Point4D by 12--14\% on the test split, against 43\% on the validation videos of Tab.~\ref{tab1}. A large part of this difference comes from the evaluation protocol. Tab.~\ref{tab1} excludes scene-exiting ground-truth tracks, whereas Tab.~\ref{tab3} follows the standard protocol and keeps them. These tracks carry an error that is large and nearly identical for every method, which dilutes any relative gain: as Tab.~\ref{tab8} shows, keeping them reduces the gain on the validation videos from 43\% to 18\%.

We therefore re-evaluate Tab.~\ref{tab3} with the filter of Sec.~\ref{secXA}, which removes 3.6\% of the tracks. The predictions are unchanged; only the set of evaluated tracks differs. As Tab.~\ref{tab10} shows, the EPE of Point4D at 2,000 frames falls from 5.334 to 1.598 and that of Recall4D from 4.634 to 0.924, so the reduction brought by Recall4D rises from 13\% to 42\%. At 1,000 and 1,500 frames it is 31\% and 40\%, in line with the 43\% on the validation videos, and it grows with video length, as expected if forgetting accumulates. SR is almost unaffected by the filter: at 1,000 frames it is 0.282 and 0.410, against 0.282 and 0.408 in Tab.~\ref{tab3}. Fewer than 4\% of the tracks thus account for most of the EPE under the standard protocol, which is why we report bounded metrics alongside it. With these tracks excluded, Point4D with Recall4D reaches the lowest EPE at every length, while TrackEverything keeps the highest APD and SR. On TrackEverything, the gain from Recall4D remains small, as explained in Sec.~\ref{secXD1}.

\subsection{A Closer Look at the All-Static Baseline}
\label{secd2}
In Tab.~\ref{tab5}, holding every point comes close to Recall4D in EPE. This baseline is the limiting case of Recall4D in which every query is labeled static: it keeps the persistent state and discards all motion. Its strength confirms our diagnosis that remembering matters more than re-predicting. It is not a tracker, however, since each of its trajectories is constant, and its accuracy depends on how static the scene is.

Tab.~\ref{tab11} compares it with Recall4D across video lengths and datasets. Besides EPE and SR over all queries, we report SR$_\text{dyn}$, the SR of dynamic queries, which is bounded and therefore not dominated by scene-exiting tracks. In EPE the two are close in most settings, since both remove the forgetting of static points, which dominates this metric. The difference lies in the dynamic queries. Holding them lowers SR$_\text{dyn}$ below that of the chained baseline in all six settings, by up to 0.119 on Dynamic Replica, whereas Recall4D stays within 0.014 of the baseline and exceeds the all-static variant by 0.03 to 0.11. Recall4D also has the highest SR over all queries in every setting. The all-static baseline thus shows how much remembering alone achieves, and Recall4D obtains this gain without giving up the dynamic part of the scene.
\begin{table*}[t]
\centering
\caption{\textbf{Design comparison with the concurrent TrackEverything~\cite{jain2026trackeverything}}, described in its default dense mode. The two methods share one idea, sparing static points from repeated trajectory prediction, and realize it by complementary routes.}
\vspace{-0.8em}
\label{tab13}
\small
\setlength{\tabcolsep}{8pt}
\begin{tabularx}{\linewidth}{@{}l >{\raggedright\arraybackslash}X >{\raggedright\arraybackslash}X@{}}
\toprule
 & TrackEverything & Recall4D (Ours)\\
\midrule
Type & Trained tracker, 41M trainable parameters & Training-free plug-in on a frozen tracker \\
Geometry & World-space pointmaps, sensed or predicted & Depth and cameras predicted by the backbone \\
Windows & Non-overlapping, 16 frames & 48 frames with stride 40, chained by Sim(3) \\
Scene state & Voxel-deduplicated world-space feature cloud & A position and a counter per static query \\
Static/dynamic split & Learned per-point classification in every window & First-frame instance masks \\
Static points & Merged by voxel, skipping trajectory refinement & Held when unseen, refined by a running mean \\
Dynamic points & Trajectory refined for every dynamic point & Up to 32 anchors per instance, rest skinned \\
\bottomrule
\end{tabularx}
\vspace{-0.5em}
\end{table*}
\begin{table*}[t]
\centering\footnotesize
\caption{\textbf{Per-sequence results of TrackEverything and Recall4D} on PointOdyssey test (200 frames). TrackEverything is run in its dense mode, and its runtime varies with the extent of the scene. The OOM column gives the video length at which the dense mode exceeds our 40\,GB memory budget.}
\vspace{-0.8em}
\label{tab14}
\setlength{\tabcolsep}{4pt}
\renewcommand{\arraystretch}{0.95}
\begin{tabular*}{\linewidth}{@{\extracolsep{\fill}} l r r ccc r c r ccc @{}}
\toprule
& & \multicolumn{6}{c}{TrackEverything~\cite{jain2026trackeverything}} & \multicolumn{4}{c}{\textbf{Recall4D (Ours)}} \\
\cmidrule(lr){3-8}\cmidrule(lr){9-12}
Sequence & \#Q & Time\,(s) & EPE$\downarrow$ & APD$\uparrow$ & SR$\uparrow$ & Geom.\,(s) & OOM & Time\,(s) & EPE$\downarrow$ & APD$\uparrow$ & SR$\uparrow$ \\
\midrule
\seqname{ani12\_new\_f}                 & 7318 &  360.8 & 1.415 & 0.328 & 0.193 & 31.2 & $\geq$1000 & 35.0 & 1.653 & 0.229 & 0.103 \\
\seqname{ani14\_new\_}                  & 8192 &  209.8 & 1.459 & 0.442 & 0.219 & 31.0 & $\geq$1000 & 28.7 & 0.779 & 0.483 & 0.229 \\
\seqname{ani2}                          & 5216 & 3031.4 & 3.372 & 0.371 & 0.230 & 31.1 & $\geq$500  & 23.8 & 1.303 & 0.159 & 0.078 \\
\seqname{anima5\_s}                     & 3857 & 2509.8 & 1.717 & 0.095 & 0.036 & 31.0 & $\geq$500  & 26.6 & 1.512 & 0.137 & 0.070 \\
\seqname{cab\_e\_3rd}                   & 8192 &   88.1 & 0.233 & 0.801 & 0.678 & 31.0 & --         & 25.7 & 0.260 & 0.806 & 0.685 \\
\seqname{dancingroom1\_3rd}             & 8192 &  162.2 & 0.761 & 0.628 & 0.606 & 30.9 & --         & 36.1 & 0.498 & 0.627 & 0.562 \\
\seqname{dancingroom1\_3rd1}            & 4808 &  324.2 & 0.233 & 0.854 & 0.818 & 30.8 & --         & 27.8 & 0.274 & 0.842 & 0.707 \\
\seqname{dancingroom1\_3rd2}            & 8192 &  275.0 & 0.867 & 0.688 & 0.660 & 30.9 & --         & 34.9 & 0.313 & 0.837 & 0.755 \\
\seqname{dancingroom3\_3rd}             & 8192 &  305.8 & 0.734 & 0.685 & 0.629 & 30.5 & --         & 34.1 & 0.479 & 0.518 & 0.394 \\
\seqname{egobody\_egocentric}           & 4354 &   62.2 & 0.122 & 0.882 & 0.777 & 31.2 & --         & 19.4 & 0.507 & 0.596 & 0.536 \\
\seqname{seminar\_g110\_0315\_3rd}      & 8192 &   88.6 & 0.143 & 0.836 & 0.826 & 30.6 & --         & 36.4 & 0.143 & 0.920 & 0.873 \\
\seqname{seminar\_g110\_0315\_ego1}     & 2903 &   69.0 & 0.061 & 0.976 & 0.973 & 31.0 & --         & 24.8 & 0.071 & 0.972 & 0.940 \\
\seqname{seminar\_g110\_0315\_ego2}     & 3191 &   63.0 & 0.093 & 0.946 & 0.922 & 31.1 & --         & 21.4 & 0.125 & 0.934 & 0.932 \\
\midrule
Mean   & & 580.8 & 0.862 & 0.656 & 0.582 & 30.9 & & 28.8 & 0.609 & 0.620 & 0.528 \\
Median & & 209.8 &       &       &       & 31.0 & & 27.8 &       &       &       \\
\bottomrule
\end{tabular*}
\vspace{-1em}
\end{table*}

\subsection{Evaluation on Real-World Videos}
\label{secXC3}
Sec.~\ref{sec61} evaluates Recall4D on two synthetic datasets, PointOdyssey~\cite{zheng2023} and Dynamic Replica~\cite{karaev2023dynamicstereo}. The reason is simple: our focus is long-horizon tracking, and existing real-world benchmarks with 3D track annotations contain only short clips. Synthetic data alone, however, cannot show whether our findings transfer to real videos. We therefore add an evaluation on the minival split of the Aria Digital Twin (ADT)~\cite{pan2023aria} subset of TAPVid-3D~\cite{koppula2024tapvid}. Among the three TAPVid-3D subsets, ADT has the longest clips, at 300 frames each, whereas DriveTrack~\cite{balasingam2024} clips range from 25 to 300 frames and Panoptic Studio~\cite{joo2015panoptic} clips have about 150. ADT is also egocentric: frequent head motion moves scene content out of view and back, which is exactly the condition under which forgetting occurs. At 300 frames, these clips remain far shorter than our long-horizon setting, so this experiment serves as a real-world validation rather than a long-horizon benchmark. As Tab.~\ref{tab12} shows, our findings transfer to real videos. Recall4D improves all six backbones that have no built-in scene memory, on every metric: EPE drops by 25\% to 45\%, SR rises by 0.06 to 0.23, and runtime falls in every case, by up to 3.9$\times$. These gains appear on clips of only 300 frames, which suggests that forgetting already matters at moderate length when the camera moves as much as a head-mounted one.

TrackEverything~\cite{jain2026trackeverything} is the most accurate method on this benchmark by a clear margin. On top of it, Recall4D leaves accuracy essentially unchanged and reduces runtime by 6\%. This matches its behavior on PointOdyssey and has the same explanation, given in Sec.~\ref{secXD1}: TrackEverything already keeps static points by design. The table thus tells one story from two sides. The tracker that was built to remember is the most accurate, and the trackers that were not move much closer to it once Recall4D lets them remember. For 4RC, the gap to TrackEverything narrows by 72\% in EPE, 79\% in APD, and 86\% in SR, without any training.
\section{Discussion and Limitations}
\label{secXD}
\subsection{A Closer Look at TrackEverything}
\label{secXD1}
Concurrent with our work, TrackEverything~\cite{jain2026trackeverything} made an impressive step for long-horizon 3D tracking: a learned tracker that follows every pixel through videos of more than a thousand frames, built on a persistent, de-duplicated 3D scene representation. As the work closest to ours, it deserves a closer look, which we give here. Tab.~\ref{tab13} compares its design with ours. Both separate static from dynamic points and spare the static ones from repeated trajectory prediction. That an independently designed and trained tracker arrives at the same decomposition is, in our view, strong support for the diagnosis of Sec.~\ref{sec3}. The two works are complementary: TrackEverything learns this behavior inside a new tracker, whereas Recall4D adds it to existing trackers without training.

\noindent\textbf{Choice of modes.} We run the released code with its default settings. It offers a dense mode, the default, which maintains the scene representation across windows, and a sparse mode, which tracks the query points without keeping the scene points of previous windows. Its authors recommend the sparse mode when the scene memory grows large and note that it has little effect on the accuracy of query points. We use each where it is intended. Tab.~\ref{tab2} compares all methods in their default configuration, so it uses the dense mode. Tab.~\ref{tab3} targets 1,000 to 2,000 frames, where the dense mode exceeds our memory budget on four of the 13 test sequences, so it uses the sparse mode. The 40\,GB budget is our own constraint, applied to every method. At 500 frames, two sequences exceed it in the dense mode, and Tab.~\ref{tab2} therefore reports OOM instead of an average over a subset. Tab.~\ref{tab14} gives the per-sequence results at 200 frames.

\noindent\textbf{Results.} TrackEverything is the strongest method in our comparison on APD and SR: it leads both on 200-frame PointOdyssey in Tab.~\ref{tab2} and at every length in Tab.~\ref{tab3}, where its EPE is also lower than that of chained Point4D. Two sequences, \texttt{ani2} and \texttt{anima5\_s}, account for most of its runtime and much of its error in Tab.~\ref{tab2}: without them, its mean tracking time is 183\,s and its mean EPE is 0.556. We stress that these two sequences are not a weakness specific to TrackEverything. Every method in our comparison, including ours, is clearly less accurate on them than on the remaining sequences, although to different degrees: Recall4D reaches an EPE of 1.303 and 1.512 on them, against 0.464 on the other eleven. The runtime of TrackEverything varies with the extent of the scene.

\noindent\textbf{Recall4D on TrackEverything.} Applied on top of TrackEverything$^\mathsf{s}$, Recall4D brings consistent but small gains: at 1,000 frames, EPE moves from 2.577 to 2.569 and SR from 0.475 to 0.485, with 15\% less runtime. This is expected. TrackEverything already classifies points as static or dynamic and does not refine the trajectories of static ones, so little forgetting is left to remove.

\begin{figure}[t]
    \centering
    \includegraphics[width=\columnwidth]{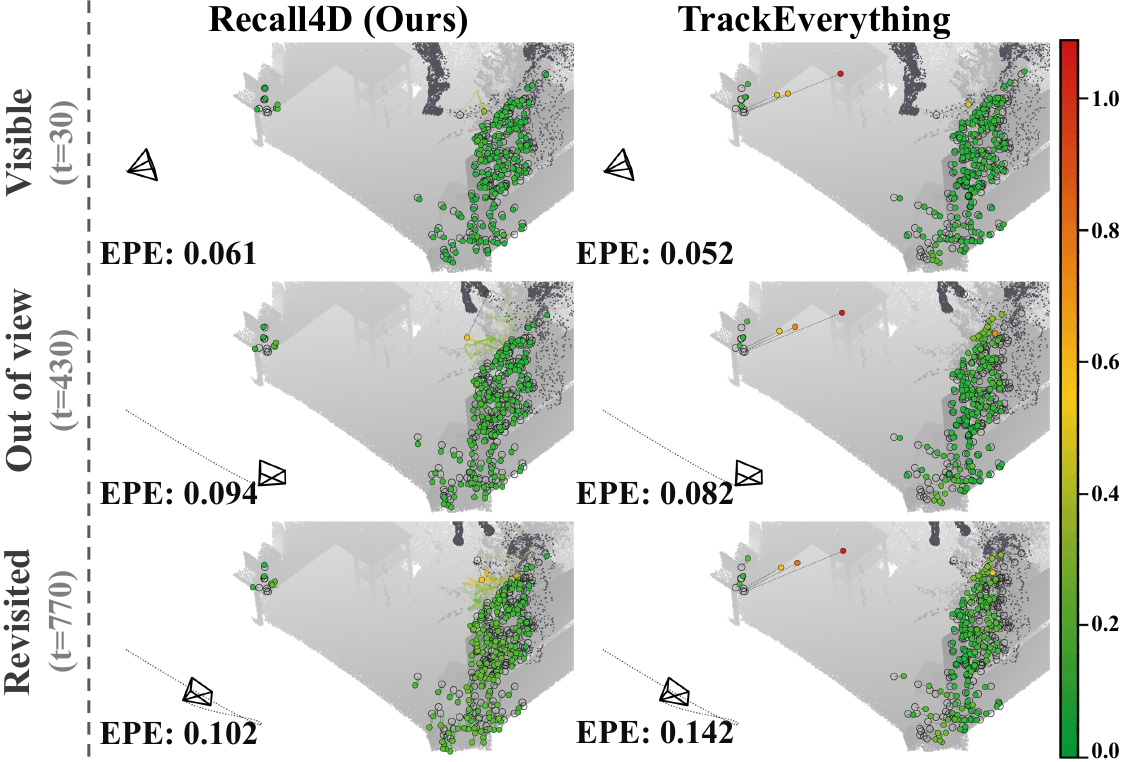}
    \vspace{-1em}
    \includegraphics[width=\columnwidth]{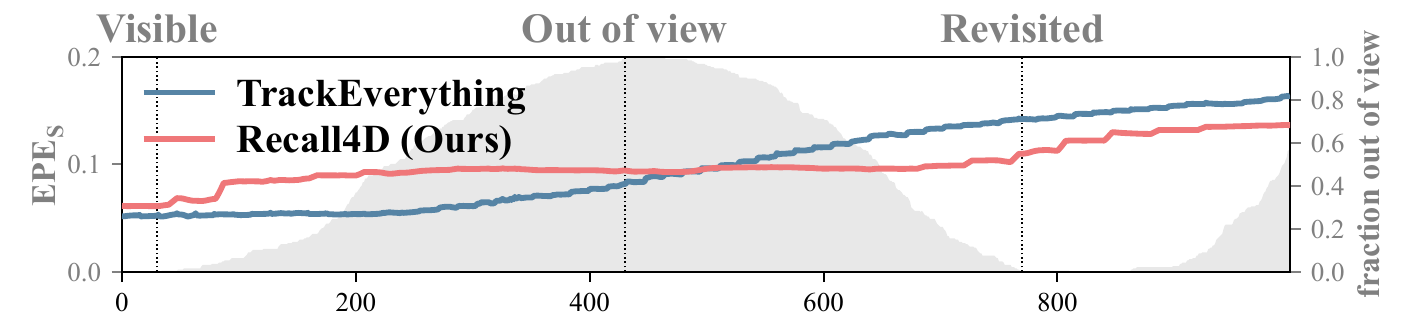}
    \vspace{-1em}
    \caption{\textbf{Static points under Recall4D and TrackEverything} on the sequence of Fig.~\ref{fig5}. Both keep static points accurate while they are out of view and after they are revisited. Bottom: EPE of static points over time; shading shows the fraction of static points outside the camera view at each frame.}
    \label{fig9}
    \vspace{-1em}
\end{figure}

\noindent\textbf{Independent evidence.} Fig.~\ref{fig9} shows the sequence of Fig.~\ref{fig5}. Static points tracked by TrackEverything stay accurate while they are out of view and after they are revisited, with an EPE of 0.082 and 0.142, against 0.411 and 0.580 for chained Point4D in Fig.~\ref{fig5}. A tracker that spares static points by design thus does not show the forgetting pattern, which is what our diagnosis predicts.

\subsection{Why Not Try to Learn Something?}
\label{secXD2}
Recall4D involves no training. We reached this design only after trying a range of alternatives, many of them learned, and none brought a measurable gain. Tab.~\ref{tab15} lists them in three groups: attempts to repair inter-window drift, attempts to handle unseen points differently, and learned variants of our components. They were run during development under settings that differ from our final protocol, so each row is compared only with its own control, and absolute values are not comparable to the tables of the main paper.
\begin{table*}[t]
\centering
\caption{\textbf{Alternatives that brought no measurable gain.} EPE on PointOdyssey val during development. Each variant is compared with its own control under identical settings. Settings differ across rows and from our final protocol, so values are not comparable to the tables of the main paper. A pair or a range gives the results of several variants of the same idea.}
\vspace{-0.8em}
\label{tab15}
\footnotesize
\setlength{\tabcolsep}{5pt}
\begin{tabularx}{\linewidth}{@{}l c c c >{\raggedright\arraybackslash}X@{}}
\toprule
Alternative & Frames & Control & Variant & Design choice it supports \\
\midrule
\multicolumn{5}{@{}l}{\emph{Repairing inter-window drift}} \\
\quad Loop closure with re-identification & 500 & 1.820 & 1.839 & Alignment is not the bottleneck \\ % TODO confirm frames
\quad Training on chained windows, last two decoder layers & 500 & 1.821 & 1.810 & Alignment is not the bottleneck \\
\quad Training on chained windows, head only & 500 & 1.821 & 1.797 & Alignment is not the bottleneck \\
\quad Latent state passed between windows & 500 & 1.800 & 1.810 & Explicit, parameter-free state \\
\quad Smoothing over overlapping frames & 500 & 1.821 & 1.821--1.823 & Alignment is not the bottleneck \\
\midrule
\multicolumn{5}{@{}l}{\emph{Handling unseen points}} \\
\quad Rigid propagation of unseen dynamic points & 500 & 1.400 & 1.399--1.401 & Dynamic motion is left to the backbone \\
\quad Offline interpolation, linear / Hermite & 500 & 1.400 & 1.405 / 1.431 & Dynamic motion is left to the backbone \\
\quad Damping of dynamic points, two strengths & 500 & 1.399 & 1.406 / 1.425 & Dynamic motion is left to the backbone \\
\quad Hold gated by decoder-predicted visibility & 500 & 1.821 & 1.695 & Visibility by projection; GT visibility reaches 1.415 \\
\midrule
\multicolumn{5}{@{}l}{\emph{Learned variants}} \\
\quad Joint fine-tuning of the decoder & 200 & 0.961 & 1.006 & Frozen backbone \\
% \quad Learned rotation head for anchors & -- & -- & -- & Displacement-only skinning; rotation error stays at $14.4^\circ$ \\
\quad Learned skinning weights with residual & 200 & 0.732 & 0.743 & Distance-based skinning \\
\quad Learned retention gate & 500 & 1.397 & 1.414 & First-frame instance masks \\ % TODO confirm frames
\quad Confidence-guided state update & 500 & 1.378 & 1.427 & Uniform running mean \\ % TODO confirm frames
\quad Fusion weighted by confidence or depth & 500 & 1.398 & 1.398 & Uniform running mean \\ % TODO confirm frames
\bottomrule
\end{tabularx}
\vspace{-1em}
\end{table*}

\noindent\textbf{Repairing inter-window drift.} Our first hypothesis was accumulated drift. Detecting revisited places with a re-identification head and correcting the chain upon a loop leaves EPE at 1.839 against 1.820. Training the tracker on chained windows, so that it sees its own handovers, changes EPE by at most 1.3\%, and passing a latent state from one window to the next does not help either. Smoothing the predicted trajectories over the overlapping frames, by linear interpolation across each overlap or by a 9-frame moving average, has no effect either: EPE stays between 1.821 and 1.823, and in the setting of Tab.~\ref{tab3} at 1,000 frames it stays within 0.001 of 2.750 for Point4D and at 2.365 for Recall4D. The windows already agree where they meet, so smoothing their junctions cannot recover what was forgotten. Together with the GT-pose alignment of Tab.~\ref{tab1}, which bounds what any alignment method can achieve, these results rule out registration as the main cause.

\noindent\textbf{Handling unseen points.} Holding helps static points but not dynamic ones, so we tried to complete the motion of unseen dynamic points: propagating them rigidly with their object, interpolating between their last and next observations, and damping their predicted motion. None improves on the control, and linear interpolation reaches only 1.349 even when both endpoints are taken from the ground truth. Recall4D therefore leaves the motion of dynamic points to the backbone and only reduces its cost. Deciding when to hold also matters: gating the hold by the decoder's predicted visibility lowers EPE by 6.9\%, against 22.3\% with ground-truth visibility. This gap led us to decide visibility by projecting the stored state into the predicted camera.

\noindent\textbf{Learned variants.} Fine-tuning the decoder jointly with our modules degraded the backbone itself, from 0.961 to 1.006 on 200-frame videos, so we keep it frozen. A learned rotation head for the anchors did not converge, its error staying at $14.4^\circ$, which suggests that the frozen decoder features carry little orientation information; our skinning therefore blends displacements only. Learned skinning weights with a residual are no better than distance-based ones, 0.743 against 0.732. A learned retention gate separates static from dynamic points better than a rule, with an AUC of 0.92 against 0.87, yet does not lower EPE. Finally, weighting observations by predicted confidence or by depth matches the uniform mean, and confidence-guided updates raise EPE from 1.378 to 1.427, since the predicted confidence is uncorrelated with the actual error. Even oracle weights would gain only 2.3\%, which is why we use a plain running mean.

% \noindent\textbf{Attempts to repair alignment.} Before the analysis of Tab.~\ref{tab1}, we tried to reduce inter-window drift directly. Loop closure leaves EPE at 1.839 against 1.820, and training on chained windows changes it by at most 1.3\%. % TODO one sentence on smoothing over the overlapping frames
% Their designs and full results are given in Sec.~\ref{secXD2}.

\subsection{Where the Speed-up Comes From}
We measure cost in decoder evaluations, where one evaluation decodes one query at one frame. The encoder runs once per window in both methods, so the saving lies entirely in the decoder. Let $N$ be the number of queries, $T$ the number of frames, and $L$ and $S$ the window length and stride, with $L{=}48$ and $S{=}40$. Chained Point4D decodes every query at every frame of each of its $T/S$ windows, which costs
\begin{equation}
C_\text{base} = \frac{T}{S}\, L\, N .
\end{equation}
Recall4D decodes the anchors at every frame, and the background only at in-view keyframes:
\begin{equation}
C_\text{ours} = \frac{T}{S}\Big(\underbrace{L \textstyle\sum_o A_o}_{\text{anchors, every frame}} \;+\; \underbrace{m\,\rho\,|\mathcal{B}|}_{\text{background, keyframes}}\Big),
\end{equation}
where $A_o$ is the number of anchors of instance $o$, $m$ the number of keyframes per window, and $\rho \le 1$ the average fraction of background points inside the camera frustum at a keyframe. The ratio of the two costs is
\begin{equation}
\frac{C_\text{ours}}{C_\text{base}} = \frac{\sum_o A_o}{N} + \frac{m}{L}\,\rho\,\frac{|\mathcal{B}|}{N}.
\end{equation}

\noindent\textbf{Static points.} A background point is decoded at $m$ of the $L$ frames, and only while it is in view. With $m{=}4$ and $L{=}48$, it costs at most one twelfth of dense decoding, and nothing once it leaves the view. Reading its position from the state at the remaining frames requires no network call, and the running-mean update is a closed-form operation.

\noindent\textbf{Dynamic points.} The decoding cost of a moving object depends on its number of anchors, at most 32, and not on the number of queries that lie on it. The remaining foreground queries are skinned: their weights over the $H{=}6$ nearest anchors are computed once, and each frame then needs only a weighted sum of anchor displacements.

\noindent\textbf{Consistency with the ablation.} The two terms act on disjoint sets of queries, so their savings should add. This matches the component ablation in Tab.~\ref{tab4}: of the 581.2\,s of chained Point4D, the persistent state alone removes 292.5\,s and the anchors alone 233.6\,s. Exactly additive savings would leave 55.1\,s, close to the measured 66.0\,s.

\noindent\textbf{Scaling.} Both costs are linear in $T$, so the speed-up is nearly constant in video length, in line with the 7.6--7.8$\times$ measured from 200 to 2,000 frames in Sec.~\ref{sec63}. In the number of queries, the two methods differ. Every added query costs chained Point4D $L$ evaluations per window. In Recall4D, an added background query costs at most $m$, and an added foreground query costs none, since it is skinned from the anchors of its instance. This is why, from 256 to 32,768 queries, the runtime of Point4D grows 46.8$\times$ and that of Recall4D only 3.4$\times$, and the speed-up reaches 13.2$\times$.

\subsection{Limitations and Outlook}
Recall4D removes the forgetting of static geometry but inherits the motion prediction of its backbone, so it does not improve how dynamic points themselves are tracked. Its decomposition relies on first-frame instance masks, and a moving object that these masks miss is treated as static. Extending memory to dynamic objects and making the decomposition adaptive over time are natural next steps.

More broadly, a feed-forward model is stateless: whatever leaves its window must be inferred again. Our results suggest that much of the long-horizon error is the price of this statelessness and not of imperfect geometry, since a small explicit state recovers a large part of it without any training. The same observation holds across five further backbones and three other tasks, and a concurrent tracker arrives at it through learning. The open problem is memory for moving objects, where a stored position no longer suffices and the state must also carry motion. 

We also believe accuracy should not be the only axis on which long-horizon 4D reconstruction is judged. A result that arrives long after the video has ended is of limited use, however accurate it is. Our experiments suggest that the two goals need not be traded against each other: removing repeated prediction lowers the error of Point4D on 1,000-frame videos while cutting its runtime from 1,176\,s to 155\,s. We see real-time long-horizon 4D reconstruction as the natural goal for the field, and we hope that future long-video models will treat what they have already seen as something to keep, and no longer as something to predict again.
\section{Extended Visualizations}
\label{secXE}
\begin{figure*}[!t]
    \centering
    \resizebox{\textwidth}{!}{\includegraphics{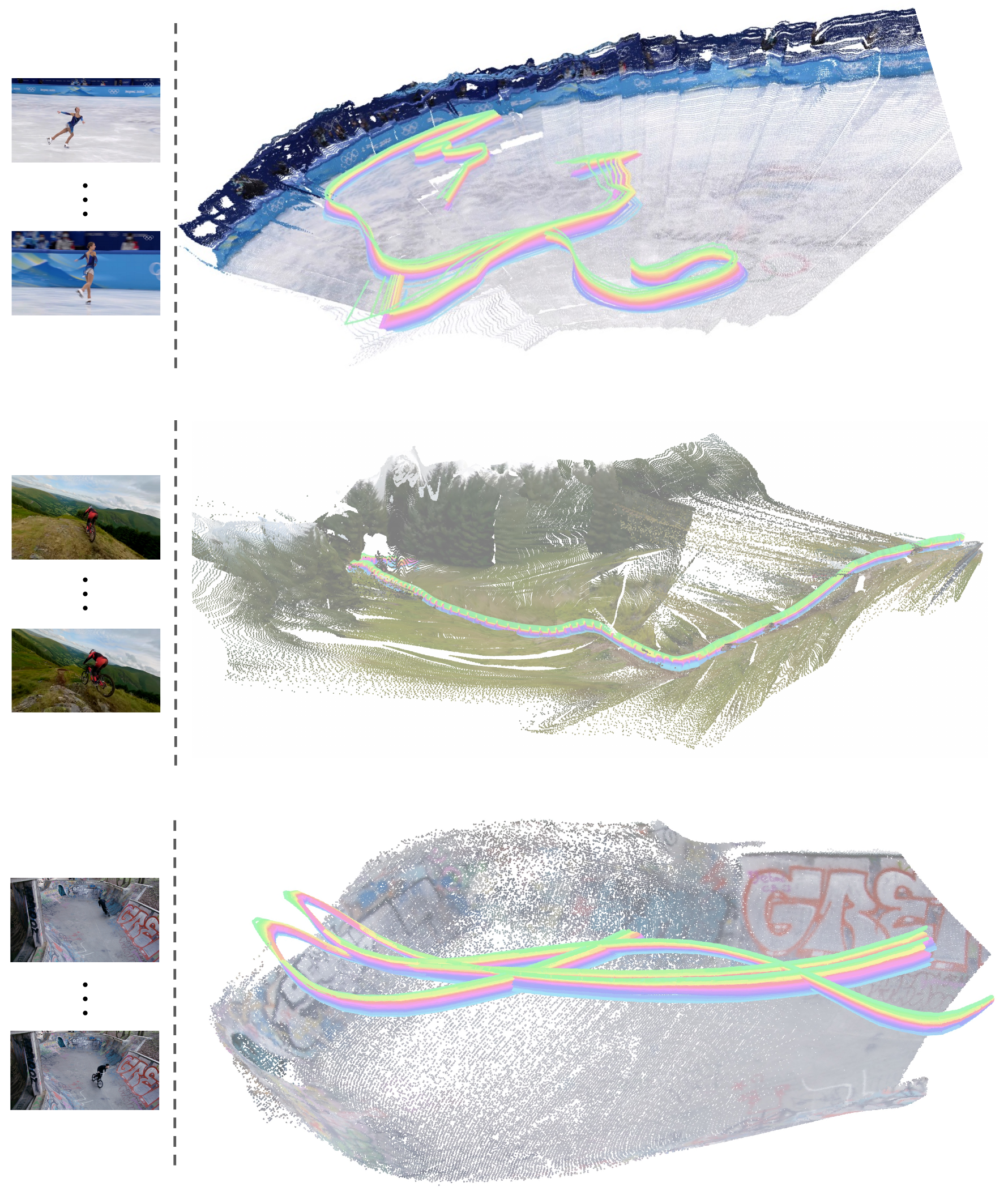}}
    \vspace{-2em}
    \caption{\textbf{Recall4D on in-the-wild videos of more than 2,000 frames.} Left: two frames of the input video. Right: the reconstructed static scene, with the trajectories of the points on the moving subject in color.}
    \vspace{-1em}
    \label{fig10}
\end{figure*}

Fig.~\ref{fig10} shows Recall4D on three in-the-wild videos with large camera and subject motion. Each video has more than 2,000 frames, and tracking takes 4\,min 27\,s per video on average. In every example, the static scene is reconstructed once and kept in the persistent world state, while the moving subject is tracked through sparse anchors.
\end{document}